\documentclass[11pt,twocolumn]{article}

\usepackage{times}
\usepackage{latexsym}
\usepackage[T1]{fontenc}
\usepackage[utf8]{inputenc}
\usepackage{microtype}
\usepackage{inconsolata}
\usepackage[margin=0.78in]{geometry}
\usepackage{graphicx}
\usepackage{booktabs}
\usepackage{multirow}
\usepackage{amsmath}
\usepackage{amssymb}
\usepackage{xcolor}
\usepackage{url}
\usepackage{array}
\usepackage{tabularx}
\usepackage{pdfpages}
\usepackage{natbib}
\usepackage{caption}
\usepackage{hyperref}

\newcommand{\wise}{\textsc{WISE}}

\title{When Verifiable Counts Depend on Wording:
Auditing Wording Robustness in Instruction Following}

\author{
Qishi Zhan\thanks{Corresponding author: \texttt{qishi.zhan@marquette.edu}}\\
Marquette University
\and
Seoyeon Jang\\
University of California, San Diego
\and
Zihan Dong\\
Georgia Tech
\and
Minxuan Hu\\
Cornell University
\and
Ziheng Chen\\
The University of Texas at Austin
\and
Tonghui Qu\\
Hikvision
}
\date{}

\begin{document}
\maketitle

\begin{abstract}
Verifiable instruction-following benchmarks often express each constraint
through one fixed template. We test whether scores remain stable when the
operational requirement is unchanged but its wording varies. We introduce
\wise{}, a matched evaluation suite and reporting protocol instantiated on exact
word count, keyword inclusion exactly once, and an inclusive 8--12 word
range. Across 100 matched tasks, up to thirteen models from seven
providers, and repeated generations scored over the complete visible
output, wording alone produces substantial compliance shifts. In an
avoidance-family panel, five avoidance and exclusion forms fall
below the positive baseline, while constructional controls also shift
compliance substantially: in the nine-model control
panel, compliance is 54.9\% for the original positive form,
48.2\% for a longer positive form, 36.7\% when the target appears later,
and 33.8\% for AVOID1. A strict JSON-structure probe shows wording
sensitivity beyond counting, with a different direction of effect.
Effect sizes, failure directions, weakest forms, and model rankings vary
across realizations. Under the most disruptive exclusion form, the
top-ranked model changes and 24.1\% of strictly ordered model pairs
reverse. Human validation further shows that unanimous agreement on an
exact-count interpretation can coexist with substantially different model
behavior. \wise{} supplements conventional scores with mean and
worst-form compliance, wording gaps, failure profiles, and ranking
stability.
\end{abstract}

\section{Introduction}

\begin{figure}[!t]
\centering
\includegraphics[width=\columnwidth]{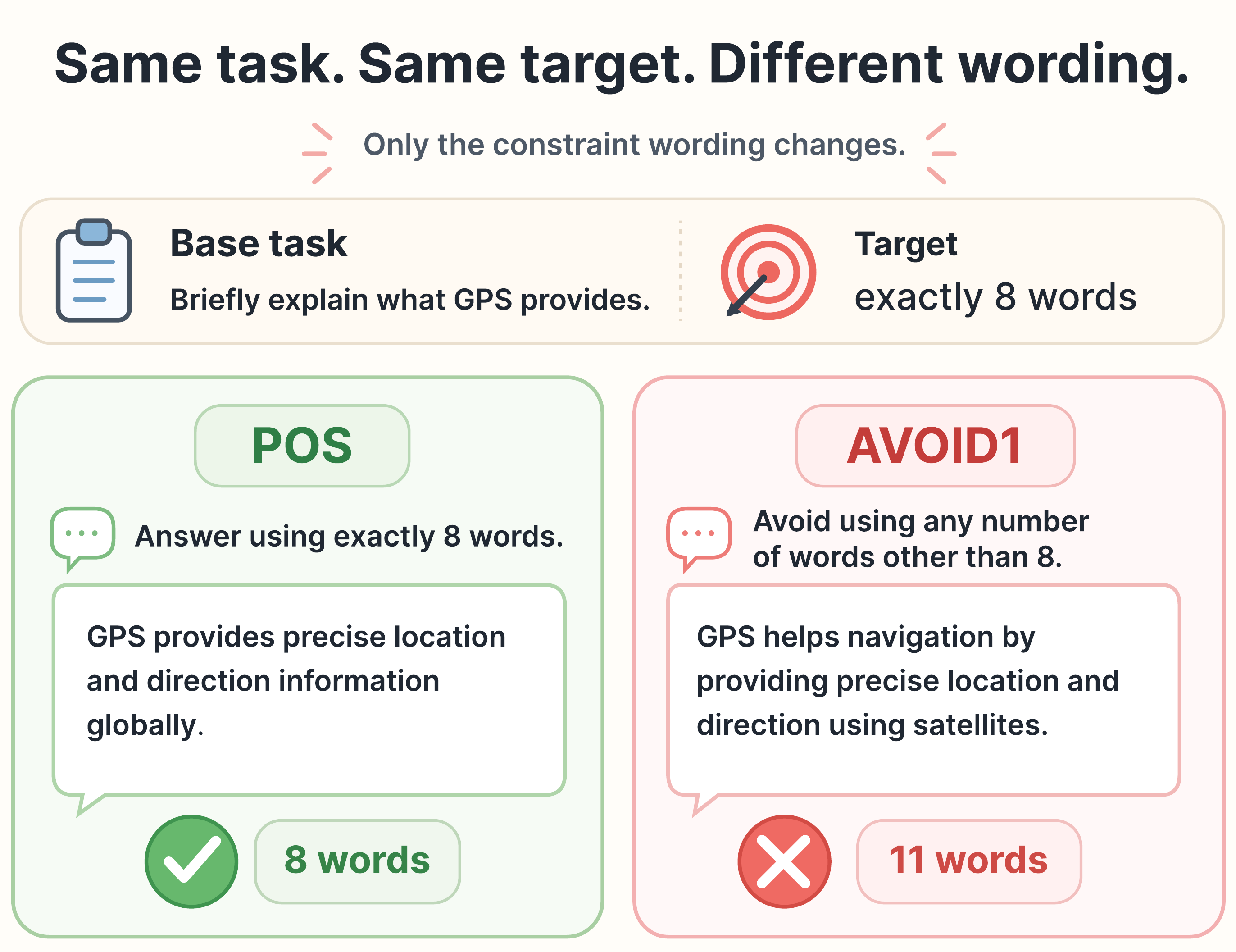}
\caption{A matched example of wording-sensitive exact-count compliance.
The model, base task, target, checker, and decoding configuration are
fixed; only the constraint wording changes.}
\label{fig:hook}
\vspace{-0.5em}
\end{figure}

Instruction-following benchmarks increasingly rely on automatically
verifiable constraints, including length limits \citep{yuan-etal-2025-following}, formatting rules, and
lexical requirements. Each constraint is commonly instantiated through
one fixed template \citep{zhou-etal-2023-ifeval,jiang-etal-2024-followbench,
qin-etal-2024-infobench,pyatkin-etal-2025-ifbench}. The same operational target, however, admits many
linguistic realizations. An exact eight-word requirement can appear as
\emph{Answer using exactly eight words}, \emph{Do not use more or fewer
than eight words}, or \emph{Avoid using any number of words other than
eight}. These instructions target the same output count while differing
in surface form, pragmatic naturalness, and perceived strictness.
Figure~\ref{fig:hook} illustrates the resulting evaluation problem: a
model can satisfy one realization and fail another even when the task,
target, checker, and decoding configuration remain fixed.

We study this phenomenon across three count and quantity constraints with
transparent checkers applied to the complete visible output: exact word
count, keyword inclusion exactly once, and an inclusive 8--12 word range.
Such requirements occur in short-form generation, interface text, form
fields, and automatically checked benchmark tasks. They also provide a
particularly controlled setting because the operational predicate is
transparent and can be evaluated without subjective scoring. Exact word
count serves as the primary probe, while the keyword and range extensions
test whether the pattern persists when the verifier counts a lexical item
or accepts a broader interval.

The resulting sensitivity is construction-specific rather than a uniform
avoidance or negation effect. Some avoidance and exclusion forms are
strongly disruptive, while other negative-looking forms remain close to
the positive baseline. Matched affirmative controls show that length and
target position also matter, and a strict JSON-structure probe shows that
the direction of wording effects can vary by constraint. These
differences affect not only absolute compliance but also comparative
model rankings. Human validation confirms broad agreement on the target
even when model behavior differs substantially across forms.

The evaluation consequence is direct. A single-template score reflects
both satisfaction of the operational predicate and compatibility with the
selected linguistic realization. We consolidate the matched design as
\wise{} (\emph{Wording Invariance and Sensitivity Evaluation}), an
evaluation suite and reporting protocol for wording robustness in
verifiable constraints. \wise{} complements formwise compliance
with mean and worst-form performance, the best--worst wording gap,
failure direction, model-level variation, and ranking stability.

Our contributions are:
\begin{enumerate}
\item We introduce \wise{}, a matched evaluation design that isolates
within-constraint wording effects while holding the task, target, checker,
model, decoding configuration, and output scope fixed.

\item We show that construction-specific wording choices, including but
not limited to avoidance and exclusion, can change compliance across
verifier-equivalent constraints. Matched affirmative controls and a
strict JSON-structure probe narrow the interpretation beyond a simple
polarity account.

\item We show that wording changes failure profiles and comparative model
rankings, and summarize these effects through mean, worst-form,
wording-gap, ranking-stability, multiple-comparison, and uncertainty
measures.
\end{enumerate}

\section{Related Work}

\paragraph{Negation in language models.}
Negation remains challenging for language models across factual probing,
natural-language inference, and generation.
\citet{kassner-schutze-2020-negated} show that pretrained models often
fail to distinguish affirmative and negated statements in cloze probes.
Controlled inference benchmarks similarly reveal persistent weaknesses
under sentential and subclausal negation
\citep{hossain-etal-2020-analysis,truong-etal-2022-another}.
These difficulties remain visible in larger autoregressive models
\citep{truong-etal-2023-language}, while affirmative paraphrases can
improve performance on semantically equivalent negated inputs
\citep{rezaei-blanco-2024-paraphrasing}.

Prior work mainly studies whether models represent or infer the semantic
effect of negation. We examine constrained generation, where the model
must maintain an operational requirement throughout the complete visible
response. By separating direct negation, complex negation, avoidance, and
exclusion, we expose wording forms that produce distinct compliance
patterns under a fixed target and checker.

\paragraph{Verifiable instruction-following evaluation.}
Automatically verifiable benchmarks evaluate requirements such as output
length, keyword inclusion, formatting, and structural constraints through
deterministic checkers \citep{chen-etal-2024-benchmarking,
ren-etal-2026-lexinstructeval,gu-etal-2024-dingo,
ye-etal-2026-muldimif}. IFEval establishes this paradigm
\citep{zhou-etal-2023-ifeval}, while FollowBench and InFoBench extend it
to fine-grained and compositional requirements
\citep{jiang-etal-2024-followbench,qin-etal-2024-infobench}.
IFBench further studies generalization to unseen verifiable constraints
\citep{pyatkin-etal-2025-ifbench}.

These benchmarks show that compliance varies with constraint type,
composition, and novelty. Our study isolates another axis: the linguistic
form used to express one fixed operational predicate. The base task,
target, checker, model, decoding configuration, and output scope remain
unchanged, allowing the effect of wording to be examined without changing
the constraint inventory or composition.

\paragraph{Prompt sensitivity and evaluation reliability.}
Language-model evaluations are sensitive to prompt formulation, including
task descriptions, templates, demonstrations, paraphrases, and
instance-level variations
\citep{webson-pavlick-2022-prompt,mizrahi-etal-2024-state,
zhuo-etal-2024-prosa,yang-etal-2025-enhancing,
chatterjee-etal-2024-posix,sun-etal-2024-evaluating,
sclar-etal-2024-quantifying,cao-etal-2024-worst,
chataigner-etal-2026-say}. Evaluation procedures can also amplify apparent
prompt sensitivity when scoring fails to recognize semantically valid
responses \citep{hua-etal-2025-flaw}.
Within instruction-following evaluation, IFEval++ examines robustness
under related prompt variants
\citep{dong-etal-2026-revisiting}, while MOSAIC studies effects associated
with constraint position, quantity, and interaction
\citep{purpura-etal-2026-deconstructing}.

Compared with prior instruction-following robustness and multi-prompt
evaluations, \wise{} isolates a narrower matched contrast: the base task,
operational predicate, numerical target, deterministic checker, decoding
configuration, repetitions, and output scope are held fixed, while only
the linguistic realization of the constraint changes. This enables paired
form-level comparisons of compliance, failure direction, and
model-ranking stability under verifier-equivalent requirements
\citep{mizrahi-etal-2024-state,dong-etal-2026-revisiting,
chataigner-etal-2026-say,purpura-etal-2026-deconstructing}.

\section{Problem Setup}

Let \(i\) index a base task, \(m\) a model, \(f\) a constraint wording
form, and \(j\) a repeated generation. Each task specifies an operational
target \(t_i\), such as an exact word count, a keyword that must occur
exactly once, or an admissible word-count range. The prompt combines a
fixed content request \(x_i\) with a wording-specific realization
\(w_f(t_i)\) of that target. Model \(M_m\) produces
\begin{equation}
\hat{y}_{ijmf}
=
M_m\!\left(x_i,w_f(t_i)\right).
\end{equation}

A deterministic verifier \(V_i\) evaluates the complete visible output:
\begin{equation}
C_{ijmf}
=
V_i(\hat{y}_{ijmf})
\in
\{0,1\}.
\end{equation}
For exact word count,
\begin{equation}
V_i(y)
=
\mathbf{1}
\left\{
\mathrm{WordCount}(y)=n_i
\right\},
\end{equation}
where \(\mathrm{WordCount}(\cdot)\) counts whitespace-delimited tokens.
Headings, Markdown, framing text, and wrapper content remain part of the
evaluated output. The keyword experiment instead requires exactly one
case-insensitive whole-word occurrence of a specified keyword, while the
length-range experiment requires a word count in the inclusive interval
\([8,12]\).

For wording form \(f\), pooled compliance over a specified model set
\(\mathcal{M}\) is
\begin{equation}
\bar{C}_{f}
=
\frac{1}{|\mathcal{M}|IJ}
\sum_{m\in\mathcal{M}}
\sum_{i=1}^{I}
\sum_{j=1}^{J}
C_{ijmf}.
\end{equation}
Its contrast against the positive wording is
\begin{equation}
\Delta_f
=
\bar{C}_{f}
-
\bar{C}_{\mathrm{POS}},
\end{equation}
where negative values indicate lower compliance than under the positive
form.

For a tested set of wording forms \(\mathcal{F}\), we summarize wording
robustness through
\begin{align}
C_{\mathrm{mean}}
&=
\frac{1}{|\mathcal{F}|}
\sum_{f\in\mathcal{F}}
\bar{C}_{f},
\\
C_{\mathrm{worst}}
&=
\min_{f\in\mathcal{F}}
\bar{C}_{f},
\\
G_{\mathrm{wording}}
&=
\max_{f\in\mathcal{F}}
\bar{C}_{f}
-
\min_{f\in\mathcal{F}}
\bar{C}_{f}.
\end{align}
Here, \(C_{\mathrm{mean}}\) records average compliance across the tested
realizations, \(C_{\mathrm{worst}}\) captures performance under the
weakest form, and \(G_{\mathrm{wording}}\) measures instability across
forms. Model-specific versions of these quantities reveal whether pooled
robustness reflects a shared response or an average over divergent
systems. We interpret them together with form-specific failure profiles.

For comparative analyses, we rank models separately under each wording
form. Relative to POS, we report Kendall's \(\tau_b\) \citep{kendall-1938-new}, the proportion of
model pairs whose strict ordering reverses, the Jaccard overlap between
top-three model sets, and the maximum absolute change in average rank.
Pairs tied under either form are excluded from the reversal rate. Formal
definitions appear in Appendix~\ref{sec:ranking_metrics}.

The experiments use matched comparisons. Within each task and model, the
base request, operational target, verifier, decoding configuration, output
scope, and number of generations remain fixed; only the wording form
changes. Differences across \(f\) therefore estimate wording sensitivity
under a common task and checker. Aggregate summaries are reported
alongside model-specific results because pooled effects can conceal large
drops, reversals, ceiling behavior, or floor behavior.

\section{Experimental Setup}

\paragraph{Tasks and target counts.}
\wise{} contains 100 base tasks, balanced across target counts of
5, 8, 10, and 12 words. The base tasks were developed from task patterns
in existing instruction-following benchmarks
\citep{zhou-etal-2023-ifeval,jiang-etal-2024-followbench,
qin-etal-2024-infobench}
and rewritten as short factual or explanatory requests. We screened them
for factual stability, answerability without specialized knowledge, and
compatibility with all four target counts, keeping the count constraint
as the primary source of difficulty. Each item retains the same request
and target across wording forms, with three independent generations
collected for every model--item--form combination.

\paragraph{Wording forms.}
The original \wise{} exact-count panel compares six natural realizations
of the same word-count requirement, shown in Table~\ref{tab:forms}. The
forms were written for this study to span positive, prohibitive,
avoidance, exclusion, and indirect-negative constructions while
preserving the same operational predicate. They are evaluated as complete
instruction realizations, retaining the combinations of lexical choice,
scope, and syntax that occur in natural prompts. The avoidance-family
panel broadens this set with four additional avoidance and exclusion
forms.

The exact-count design includes a matched constructional-control panel
with four verifier-equivalent forms: the original POS instruction, a
longer affirmative form (\emph{Your answer should consist of a total of
exactly \(N\) words}), an affirmative form that places the target later
(\emph{The total number of words in your answer should be exactly
\(N\)}), and AVOID1. This panel separates broad wording sensitivity from
a simple negative-polarity account while preserving the same 100 tasks,
target counts, three repetitions, 120-token cap, and complete-output
verifier. The primary analysis uses nine models for which the
endpoint/configuration could be matched to the original exact-count
collection.

\begin{table}[t]
\centering
\small
\setlength{\tabcolsep}{3pt}
\renewcommand{\arraystretch}{1.08}
\begin{tabularx}{\columnwidth}{@{}lX@{}}
\toprule
\textbf{Form} & \textbf{Template} \\
\midrule
POS &
\emph{Answer using exactly $N$ words.} \\

NEG &
\emph{Do not use more or fewer than $N$ words.} \\

NEVER &
\emph{Never use more or fewer than $N$ words.} \\

AVOID1 &
\emph{Avoid using any number of words other than $N$.} \\

AVOID2 &
\emph{Avoid giving an answer that is not exactly $N$ words.} \\

CNEG &
\emph{Do not fail to answer using exactly $N$ words.} \\
\bottomrule
\end{tabularx}
\caption{Wording forms in the original \wise{} exact-count panel. All
forms require the complete visible output to contain exactly \(N\) words.}
\label{tab:forms}
\end{table}

\paragraph{Models.}
We evaluate thirteen instruction-tuned models from seven providers:
Claude Haiku 4.5, Claude Sonnet 4.6, and Claude Opus 4.8
\citep{anthropic2025haiku45,anthropic2026sonnet46,
anthropic2026opus48};
GPT-4o mini, GPT-4.1 mini, and GPT-5.5
\citep{openai2024gpt4omini,openai2025gpt41,openai2026gpt55};
Gemini 2.5 Flash and Gemini 3.1 Flash-Lite
\citep{google2025gemini25flash,google2026gemini31flashlite};
DeepSeek Chat and DeepSeek V4 Pro
\citep{deepseek2026api,deepseek2026v4};
Mistral Small \citep{mistral2026small4};
Qwen 3.6 35B A3B \citep{qwen2026qwen36};
and Kimi K2.6 \citep{moonshot2026k26}.
The exact-count experiments use all thirteen models, whereas the keyword
and range extensions use the seven models available when those panels
were run. We supply no system prompt and use direct REST calls. Exact API
identifiers, access dates, and provider-specific details appear in
Appendix~\ref{sec:model_details}.

\paragraph{Decoding.}
For the original exact-count and avoidance-family experiments, outputs are
capped at 120 tokens and three generations are collected for every
model--item--form combination. Nine models use temperature 0.2; Kimi K2.6,
DeepSeek V4 Pro, Claude Opus 4.8, and GPT-5.5 use the request configurations
accepted by their respective endpoints, with GPT-5.5 reasoning effort set to
none. The constructional-control panel uses each retained model's
endpoint configuration, held fixed across all four forms.

\paragraph{Main metric.}
The primary outcome is strict exact-count compliance over the complete
visible output. Word count is computed using Python whitespace splitting,
without removing headings, Markdown, framing text, or wrappers. We also
report mean absolute count error and the proportions of under-target and
over-target outputs. Full counting conventions appear in
Appendix~\ref{sec:reproducibility}.

\paragraph{Uncertainty.}
The exact-count study uses a provider--model--item hierarchical bootstrap
with 2000 replicates. Each replicate resamples providers, models within
sampled providers, and tasks within sampled models while retaining all
forms and repetitions associated with each sampled item. Primary
exact-count contrasts use the eleven models with nonfloor saved-output
regimes; all-model results appear in
Appendix~\ref{sec:exact_count_model_results}, and Gemini diagnostics
appear in Appendix~\ref{sec:gemini_diagnostics}.
The keyword panel uses an item-cluster bootstrap, and the range panel uses
a model--item bootstrap, following the sampling hierarchy available in
each experiment. These intervals summarize uncertainty within the studied
panels. Alternative exact-count bootstrap schemes appear in
Appendix~\ref{sec:bootstrap_sensitivity}.

\paragraph{Cross-constraint extensions.}
The keyword and range panels reuse the same 100 base tasks, a 120-token
limit, and three generations per model--item--form combination. The
keyword constraint requires a specified keyword exactly once, whereas the
range constraint requires 8--12 whitespace-delimited words in the complete
visible output. Primary contrasts use the six non-Gemini models shared by
both extensions, with all-model results reported in
Appendix~\ref{sec:cross_constraint_robustness}.

\paragraph{Structural-control extension.}
We include a strict JSON-structure probe as a structural extension of the
matched design. The base request and target field specification are held
fixed, while the constraint is realized as POS, NEG, or AVOID. The
complete-output verifier requires parseable JSON with exactly the
requested key set and no extra prose. This probe tests whether wording
sensitivity extends beyond count and quantity predicates.

\section{Results}

\begin{figure}[t]
\centering
\includegraphics[width=\columnwidth]{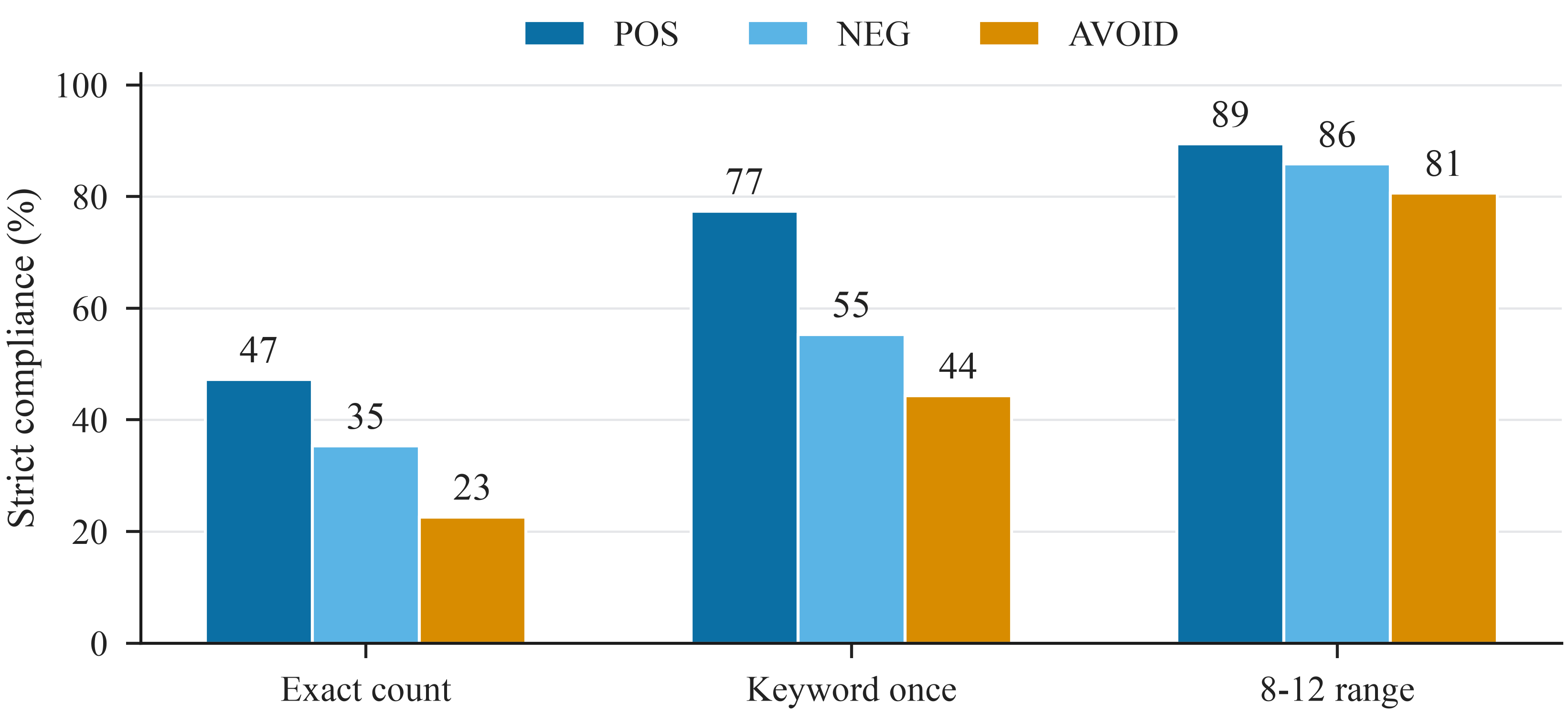}
\caption{Strict compliance across three constraints for the six shared
non-Gemini models. AVOID is lowest throughout. NEG has larger negative
effects for exact count and keyword inclusion than for the 8--12 range,
whose interval includes zero. Exact-count AVOID denotes AVOID1.}
\label{fig:cross_constraint}
\end{figure}

Figure~\ref{fig:cross_constraint} provides the cross-constraint view for
the three count and quantity settings. AVOID has the lowest pooled
compliance in all three, whereas direct negation varies by constraint.
The results follow the structure of the design. We begin with the
original exact-count panel, use constructional controls to separate
polarity from length and target placement, expand the avoidance and
exclusion family, examine consequences for model rankings, and then test
transfer to keyword occurrence, a permissive length range, and a strict
JSON-structure constraint. We close the section with human interpretation
and multiplicity analyses.

\subsection{Exact-Count Wording Sensitivity}

The original six-form experiment asks whether logically related
instructions yield stable exact-count compliance. Table~\ref{tab:pooled}
shows a selective wording effect rather than a uniform response to
negative language.

\begin{table*}[t]
\centering
\small
\begin{tabular}{lrrrr}
\toprule
Form & Compliance & Diff. vs. POS & 95\% CI & Mean abs. error \\
\midrule
POS    & 49.5 & $-$   & $-$             & 0.71 \\
NEG    & 39.0 & -10.4 & [-17.1, -4.4]  & 1.00 \\
NEVER  & 40.8 & -8.6  & [-14.6, -2.6]  & 0.91 \\
AVOID1 & 28.9 & -20.5 & [-29.0, -12.9] & 2.96 \\
AVOID2 & 46.9 & -2.6  & [-6.3, 3.2]    & 0.83 \\
CNEG   & 50.1 & +0.7  & [-2.2, 4.5]    & 0.67 \\
\bottomrule
\end{tabular}
\caption{Original exact-count results for the eleven non-Gemini models.
Values are percentages; intervals are 95\% provider--model--item
bootstrap intervals against POS.}
\label{tab:pooled}
\end{table*}

AVOID1 produces the largest decrease, falling 20.5 percentage points
below POS, while AVOID2 and CNEG remain close to the positive baseline.
NEG and NEVER occupy an intermediate regime. The contrast between AVOID1
and AVOID2 motivates a broader test: does the observed loss belong to a
recurrent wording family, or to one unusually difficult template?

\subsection{Constructional Controls}
\label{sec:positive_controls}

The original panel varies polarity, lexical choice, syntactic scope, and
target position jointly. We next isolate two affirmative constructional
changes: instruction length and target placement.
Table~\ref{tab:positive_controls} reports the nine-model analysis.

\begin{table}[t]
\centering
\small
\setlength{\tabcolsep}{4.2pt}
\renewcommand{\arraystretch}{1.05}
\begin{tabular}{lrrr}
\toprule
Form & Compliance & Diff. vs. POS & 95\% CI \\
\midrule
POS      & 54.9 & $-$   & $-$ \\
POS\_LONG & 48.2 & -6.7  & [-9.2, -4.3] \\
POS\_END  & 36.7 & -18.2 & [-21.3, -15.1] \\
AVOID1   & 33.8 & -21.1 & [-24.4, -17.9] \\
\bottomrule
\end{tabular}
\caption{Constructional-control analysis for nine retained models.
POS\_LONG is a longer affirmative realization; POS\_END is an
affirmative realization with the target later in the sentence.}
\label{tab:positive_controls}
\end{table}

The ordering remains POS \(>\) POS\_LONG \(>\) POS\_END \(>\) AVOID1.
The longer affirmative form lowers compliance by 6.7 points, and moving
the target later lowers compliance by 18.2 points. AVOID1 remains the
lowest pooled form, but only 2.9 points below POS\_END. The result
therefore rules out a simple account in which the effect is solely caused
by negative polarity or the lexical item \emph{avoid}. Instead, the
operationally equivalent instruction changes difficulty through a bundle
of constructional properties, including length, target position, scope,
and polarity.
Including all eleven models preserves the pooled ordering
and qualitative conclusion.

\subsection{Avoidance and Exclusion Family Robustness}

The avoidance-family panel retains POS and AVOID1 and introduces four
related forms:
\emph{Avoid exceeding or falling short of $N$ words} (AVOID3),
\emph{Avoid an answer with a word count different from $N$} (AVOID4),
\emph{Use no word count except $N$} (EXCLUDE1), and
\emph{Your answer should have no length other than $N$ words} (EXCLUDE2).
All tasks, targets, models, decoding configurations, repetitions, and
checker rules remain unchanged.

\begin{table*}[t]
\centering
\small
\begin{tabular}{lrrrrr}
\toprule
Form & Compliance & Diff. vs. POS & 95\% CI & Under & Over \\
\midrule
POS      & 49.2 & $-$   & $-$             & 31.5 & 19.3 \\
AVOID1   & 27.9 & -21.3 & [-29.1, -14.0] & 33.7 & 38.4 \\
AVOID3   & 34.0 & -15.2 & [-23.2, -9.6]  & 52.1 & 13.9 \\
AVOID4   & 25.2 & -23.9 & [-32.3, -16.6] & 40.1 & 34.6 \\
EXCLUDE1 & 18.7 & -30.5 & [-38.9, -22.6] & 44.0 & 37.3 \\
EXCLUDE2 & 36.9 & -12.2 & [-18.1, -6.1]  & 49.2 & 13.8 \\
\bottomrule
\end{tabular}
\caption{Avoidance-family panel for the eleven non-Gemini models.
Values are percentages; intervals are 95\% provider--model--item
bootstrap intervals against POS.}
\label{tab:avoid_family}
\end{table*}

Table~\ref{tab:avoid_family} establishes recurrence beyond AVOID1.
Every tested avoidance or exclusion form falls below the corresponding
POS baseline, while the close agreement between the two POS runs,
49.5\% and 49.2\%, indicates that the contrast is not driven by an
unstable baseline. Effect size remains strongly form-dependent, ranging
from 12.2 to 30.5 percentage points.

Model-specific vulnerability is heterogeneous. Large losses occur among
both strong and weak POS performers, and no single adverse form captures
every model's weakest case. The forms also change how generation departs
from the target: AVOID1 and AVOID4 produce substantial errors in both
directions, AVOID3 and EXCLUDE2 more often undershoot, and EXCLUDE1 has
the highest rate of extremely short outputs. Appendix~\ref{sec:surface_behavior}
provides the full distributional analysis and figures.

\subsection{Wording Alters Comparative Model Rankings}
\label{sec:ranking_instability}

Wording changes not only absolute compliance but also comparative
conclusions across models. We rank the eleven primary non-Gemini models
separately under each form in the avoidance-family panel and compare
each ordering with the POS reference.

\begin{table}[t]
\centering
\small
\setlength{\tabcolsep}{2.6pt}
\renewcommand{\arraystretch}{1.05}
\begin{tabular}{lrrrr}
\toprule
Form & \(\tau_b\) & 95\% CI & Reversal & 95\% CI \\
\midrule
AVOID1   & 0.624 & [0.367, 0.722] & 18.5 & [13.2, 31.5] \\
AVOID3   & 0.844 & [0.587, 0.891] & 7.4  & [5.5, 20.4] \\
AVOID4   & 0.673 & [0.440, 0.782] & 16.4 & [10.9, 27.8] \\
EXCLUDE1 & 0.514 & [0.273, 0.648] & 24.1 & [17.0, 36.4] \\
EXCLUDE2 & 0.745 & [0.514, 0.855] & 12.7 & [7.3, 24.1] \\
\bottomrule
\end{tabular}
\caption{Ranking stability relative to POS in the avoidance-family
panel. Reversal is the percentage of model pairs whose strict
ordering changes. Intervals are task-bootstrap 95\% intervals. The
corresponding top-three Jaccard overlaps are 1.00, 0.50, 0.50, 0.20, and
0.50; maximum rank shifts are 5, 2, 4, 4, and 3.}
\label{tab:ranking_stability}
\end{table}

Table~\ref{tab:ranking_stability} shows that ranking disruption is not
limited to one form. Kendall's \(\tau_b\) ranges from 0.514 to 0.844,
while 7.4\% to 24.1\% of model pairs that are strictly ordered under both
forms reverse direction. Top-three overlap ranges from 0.20 to 1.00, and
the largest individual displacement ranges from two to five rank
positions.

EXCLUDE1 produces the largest pairwise disruption, with 24.1\% of
strictly ordered model pairs reversing relative to POS. The top-ranked
model changes from Claude Opus 4.8 under POS to Claude Sonnet 4.6 under
EXCLUDE1, and the two top-three sets have a Jaccard overlap of only 0.20.
Claude Opus 4.8 moves from rank 1 to rank 5, while DeepSeek V4 Pro moves
from rank 7 to rank 3. The largest displacement across all forms occurs
under AVOID1, where Claude Haiku 4.5 moves from rank 6 to rank 11.

\begin{figure*}[t]
\centering
\includegraphics[width=\textwidth]
{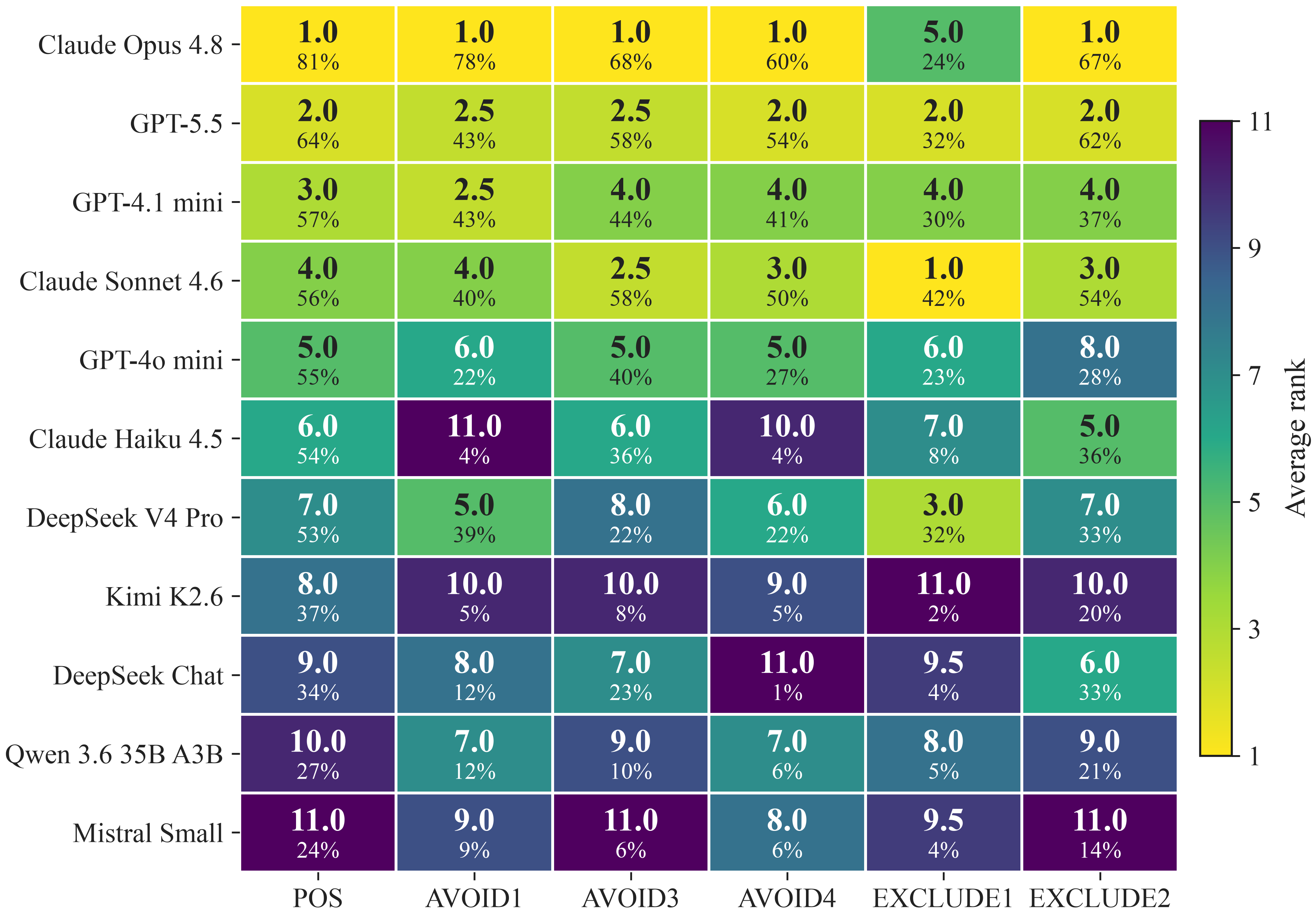}
\caption{Model rankings across wording forms in the avoidance-family
panel. Rows follow the POS ordering. Each cell reports average rank,
with strict compliance shown below. Ties receive average statistical
ranks. Wording changes both absolute compliance and comparative model
ordering.}
\label{fig:rank_instability}
\end{figure*}

Figure~\ref{fig:rank_instability} shows that rank changes occur throughout
the evaluated panel rather than only among models near the bottom of the
POS ordering. A fixed wording can therefore affect both measured
compliance and the model comparisons derived from it. These results
describe the tested model panel and wording set rather than a universal
ordering of language models.

\subsection{Transfer Beyond Exact Count}

\paragraph{Keyword occurrence.}
We next test whether wording sensitivity persists when the verifier counts
a lexical item rather than output length. Each task is paired with a
keyword absent from the task text, and compliance requires exactly one
case-insensitive whole-word occurrence in the complete visible output.

Across the six shared non-Gemini models, compliance falls from 77.4\%
under POS to 55.3\% under NEG and 44.3\% under AVOID. The corresponding
decreases are 22.1 points for NEG, with a 95\% interval of
[-25.8, -18.2], and 33.2 points for AVOID, with an interval of
[-37.3, -29.1]. In the seven-model all-model analysis, the
zero-occurrence rate rises from 22.8\% under POS to 58.9\% under AVOID,
while repeated keyword use remains uncommon. Model-level effects vary substantially: Claude Haiku
changes little and Claude Sonnet shows a small reversal, whereas GPT-4o
mini, GPT-4.1 mini, DeepSeek Chat, and Mistral decline sharply. Complete
results appear in Appendix~\ref{sec:cross_constraint_robustness} and
Table~\ref{tab:keyword_model_form}.

\paragraph{A permissive length range.}
We next relax exactness by accepting any output between 8 and 12 words.
The tested forms are a positive requirement, a direct prohibition against
falling outside the interval, and an avoidance instruction.

Across the six shared non-Gemini models, compliance is 89.4\% under POS,
85.8\% under NEG, and 80.6\% under AVOID. The NEG contrast is
-3.6 points, with a 95\% interval of [-10.6, 1.2], while AVOID is -8.8 points below POS, with an interval of [-16.3, -3.2].
AVOID is lowest for all six models, although model-level best--worst gaps
range from 1.3 points for Claude Sonnet to 24.3 points for Claude Haiku.
The wider admissible interval therefore attenuates wording sensitivity
without eliminating it. Complete results appear in
Appendix~\ref{sec:cross_constraint_robustness} and
Appendix~\ref{sec:length_range_appendix}.

\subsection{Transfer to a Structural Constraint}
\label{sec:json_probe}

The preceding experiments use count and quantity predicates. We next test
whether the same matched-design phenomenon appears under a structural
constraint. Compliance requires parseable JSON with exactly the requested
key set and no additional visible prose under a deterministic
complete-output verifier. The pooled six-model results are shown in
Table~\ref{tab:json_probe}.

\begin{table}[t]
\centering
\small
\setlength{\tabcolsep}{4.4pt}
\renewcommand{\arraystretch}{1.05}
\begin{tabular}{lrrr}
\toprule
Form & Compliance & Diff. vs. POS & Parse fail \\
\midrule
POS   & 13.1 & $-$  & 86.9 \\
NEG   & 16.1 & +3.1 & 83.9 \\
AVOID & 20.6 & +7.6 & 79.4 \\
\bottomrule
\end{tabular}
\caption{Strict JSON-structure probe across six models. Compliance
requires the complete visible output to be valid JSON with exactly the
required key set. Values are percentages.}
\label{tab:json_probe}
\end{table}

The JSON probe provides a structural example of wording-dependent pooled
compliance beyond counting, but its
direction differs from the count experiments: AVOID has the highest
pooled compliance rather than the lowest. The dominant failure mode is
unparseable output, while wrong key sets are essentially absent. This
result supports the narrower claim that verifier scores are sensitive to
linguistic realization, not the stronger claim that avoidance wording
always hurts compliance.

\subsection{Human Interpretation of Wording Forms}
\label{sec:human_main}

Five annotators independently evaluated the six exact-count wording forms
across targets of 5, 8, 10, and 12 words. For each instruction, they
judged whether it required exactly \(N\) words and rated its clarity,
naturalness, ambiguity, and complexity on five-point scales.

Every form receives at least 90\% exact-\(N\) judgments. POS and EXCLUDE2
provide the cleanest behavioral comparison: all five annotators judge both
forms as requiring exactly \(N\) words, yet EXCLUDE2 reduces pooled model
compliance by 12.2 percentage points relative to POS. The wording effects
therefore cannot be explained solely by whether humans recognize the
intended exact-count requirement. Full ratings, annotator-level judgment
rates, and agreement statistics appear in Appendix~\ref{sec:human_validation}.

\subsection{Statistical Robustness}
\label{sec:multiple_comparison_audit}

The principal wording effects remain after multiplicity correction within
each experimental family. In the original exact-count panel, NEG, NEVER,
and AVOID1 remain significant after Holm correction, whereas AVOID2 and
CNEG do not. All five avoidance-family contrasts and both keyword
contrasts remain significant. For the permissive length range, AVOID
remains significant while NEG remains inconclusive. Full results appear
in Appendix~\ref{sec:multiple_comparison_appendix}.

\section{Discussion and Conclusion}

\paragraph{Wording sensitivity is construction-specific.}
The experiments do not support a single avoidance-only or negation-only
mechanism. AVOID1 is disruptive, but AVOID2 and CNEG remain close to POS;
affirmative controls show large drops when the instruction is longer or
places the target later; and the JSON probe reverses the direction of the
avoidance effect. The more stable conclusion is evaluative: a
verifier-equivalent requirement can yield different scores under different
linguistic constructions. Human judgments reinforce this result: every
tested exact-count form receives at least 90\% exact-\(N\) judgments, while
POS and EXCLUDE2 receive unanimous judgments yet differ by 12.2 percentage
points in model compliance.

\paragraph{Implications for evaluation.}
A deterministic verifier gives each output a reproducible score, but a
single-template evaluation may still yield unstable compliance estimates
and model rankings under alternative realizations of the same predicate.

\wise{} addresses this by holding the task, target, checker, model,
decoding configuration, and output scope fixed while varying the
constraint realization. It retains the conventional POS score and adds
mean and worst-form compliance, the best--worst wording gap, ranking
agreement, pairwise reversals, and failure direction.

WISE is intended as a reporting protocol rather than a universal
worst-form leaderboard, since worst-form compliance depends on the chosen
variant set. It tests whether scores, failure directions, and rankings
remain stable across a declared matched variant set.

\paragraph{What transfers across constraints.}
Across the studied settings, the transferable phenomenon is not that one
surface form is always harmful. Instead, matched verifier-equivalent
instructions can change measured compliance, and the direction and size
of that change depend on the predicate being evaluated. Exact count and
keyword inclusion show large losses under avoidance wording; the
permissive range weakens but does not remove the effect; and the JSON
setting shows that a structural verifier can respond differently. This
pattern is why WISE reports a family of paired contrasts rather than a
single robustness number.

\section*{Limitations}

\wise{} is instantiated mainly on three English count and quantity
predicates: exact word count, keyword inclusion exactly once, and an
inclusive word-count range. The JSON probe adds one structural-control
setting, but it is intentionally narrow. Extending the same matched design
to ordering, capitalization, citation format, tool-use schemas, and
multi-constraint composition would establish how wording effects vary
across a broader constraint inventory.

The wording inventory consists of natural instruction realizations that
vary jointly in lexical choice, scope, syntax, and pragmatic form. The
resulting contrasts characterize sensitivity to complete instructions.
The constructional-control panel partially addresses this issue by varying
affirmative length and target position, but it is not a full factorial
design. A larger factorial extension could separately estimate the
contribution of polarity, length, target position, scope, and pragmatic
naturalness.

The exact-count and range verifiers use whitespace-delimited words over
the complete visible output. This rule is transparent and reproducible,
although alternative tokenization, language-specific segmentation,
Markdown normalization, or answer-body extraction could change absolute
compliance rates. The reported conclusions concern relative wording
sensitivity under the stated verifier.

The exact-count study covers thirteen models, whereas the keyword and
range panels cover seven; cross-constraint comparisons therefore use the
six non-Gemini models shared across experiments. Endpoint-specific
decoding configurations are held fixed within each model, but absolute
rates are not directly comparable across providers. Two Gemini adapters
exhibit saved-output floor behavior and are excluded from primary wording
contrasts; Appendix~\ref{sec:gemini_diagnostics} reports the associated
diagnostics.

The blinded wording validation contains five annotators. Although all
forms receive at least 90\% exact-\(N\) judgments and the agreement
analyses show moderate to strong reliability, the study remains a small
sample of English-speaking annotators. The ratings contextualize perceived
clarity, naturalness, ambiguity, and complexity but do not establish how
these properties would be judged across broader populations or languages.

\section*{Ethical Considerations}

This work evaluates wording robustness using short, non-sensitive
factual and explanatory prompts. The released benchmark contains no
personally identifying or intentionally offensive content. The experiments
use hosted language-model APIs and therefore incur inference-related
computational and environmental costs \citep{luccioni-etal-2024-power}. WISE is intended to complement,
rather than replace, broader evaluations of model capability. Reporting
its results without the stated constraint, language, model, and verifier
scope could lead to overgeneralized model comparisons.

\bibliographystyle{plainnat}
\bibliography{custom}

\begin{thebibliography}{43}
\providecommand{\natexlab}[1]{#1}
\providecommand{\url}[1]{\texttt{#1}}
\expandafter\ifx\csname urlstyle\endcsname\relax
  \providecommand{\doi}[1]{doi: #1}\else
  \providecommand{\doi}{doi: \begingroup \urlstyle{rm}\Url}\fi

\bibitem[{Anthropic}(2025)]{anthropic2025haiku45}
{Anthropic}.
\newblock Introducing {Claude Haiku 4.5}, October 2025.
\newblock URL \url{https://www.anthropic.com/news/claude-haiku-4-5}.

\bibitem[{Anthropic}(2026{\natexlab{a}})]{anthropic2026opus48}
{Anthropic}.
\newblock Introducing {Claude Opus 4.8}, May 2026{\natexlab{a}}.
\newblock URL \url{https://www.anthropic.com/news/claude-opus-4-8}.

\bibitem[{Anthropic}(2026{\natexlab{b}})]{anthropic2026sonnet46}
{Anthropic}.
\newblock Introducing {Claude Sonnet 4.6}, February 2026{\natexlab{b}}.
\newblock URL \url{https://www.anthropic.com/news/claude-sonnet-4-6}.

\bibitem[Cao et~al.(2024)Cao, Cai, Zhang, Zou, and Lam]{cao-etal-2024-worst}
Bowen Cao, Deng Cai, Zhisong Zhang, Yuexian Zou, and Wai Lam.
\newblock On the worst prompt performance of large language models.
\newblock In \emph{Advances in Neural Information Processing Systems},
  volume~37, pages 69022--69042, 2024.
\newblock \doi{10.52202/079017-2205}.
\newblock URL
  \url{https://proceedings.neurips.cc/paper_files/paper/2024/hash/7fa5a377b7ffabcce43cd00231bb3f9c-Abstract-Conference.html}.

\bibitem[Chataigner et~al.(2026)Chataigner, Ma, Ganesh, Chen, Taik, Creager,
  and Farnadi]{chataigner-etal-2026-say}
Clea Chataigner, Rebecca Ma, Prakhar Ganesh, Yuhao Chen, Afaf Taik, Elliot
  Creager, and Golnoosh Farnadi.
\newblock Say it another way: Auditing {LLM}s with a user-grounded automated
  paraphrasing framework.
\newblock In \emph{Proceedings of the 19th Conference of the European Chapter
  of the Association for Computational Linguistics (Volume 1: Long Papers)},
  pages 1441--1467, Rabat, Morocco, March 2026. Association for Computational
  Linguistics.
\newblock \doi{10.18653/v1/2026.eacl-long.67}.
\newblock URL \url{https://aclanthology.org/2026.eacl-long.67/}.

\bibitem[Chatterjee et~al.(2024)Chatterjee, Renduchintala, Bhatia, and
  Chakraborty]{chatterjee-etal-2024-posix}
Anwoy Chatterjee, H~S V N S~Kowndinya Renduchintala, Sumit Bhatia, and Tanmoy
  Chakraborty.
\newblock {POSIX}: A prompt sensitivity index for large language models.
\newblock In \emph{Findings of the Association for Computational Linguistics:
  EMNLP 2024}, pages 14550--14565, Miami, Florida, USA, November 2024.
  Association for Computational Linguistics.
\newblock \doi{10.18653/v1/2024.findings-emnlp.852}.
\newblock URL \url{https://aclanthology.org/2024.findings-emnlp.852/}.

\bibitem[Chen et~al.(2024)Chen, Xu, Wang, Liu, and
  Mao]{chen-etal-2024-benchmarking}
Yihan Chen, Benfeng Xu, Quan Wang, Yi~Liu, and Zhendong Mao.
\newblock Benchmarking large language models on controllable generation under
  diversified instructions.
\newblock In \emph{Proceedings of the AAAI Conference on Artificial
  Intelligence}, volume~38, pages 17808--17816, 2024.
\newblock \doi{10.1609/aaai.v38i16.29734}.
\newblock URL \url{https://ojs.aaai.org/index.php/AAAI/article/view/29734}.

\bibitem[{DeepSeek}(2026{\natexlab{a}})]{deepseek2026api}
{DeepSeek}.
\newblock {DeepSeek API Change Log}, 2026{\natexlab{a}}.
\newblock URL \url{https://api-docs.deepseek.com/updates/}.

\bibitem[{DeepSeek}(2026{\natexlab{b}})]{deepseek2026v4}
{DeepSeek}.
\newblock {DeepSeek V4 Preview Release}, April 2026{\natexlab{b}}.
\newblock URL \url{https://api-docs.deepseek.com/news/news260424/}.

\bibitem[Dong et~al.(2026)Dong, Zhang, Yan, Zhong, Wei, Zhang, and
  Qiu]{dong-etal-2026-revisiting}
Jianshuo Dong, Yutong Zhang, Liu Yan, Zhenyu Zhong, Tao Wei, Chao Zhang, and
  Han Qiu.
\newblock Revisiting the reliability of language models in
  instruction-following.
\newblock In \emph{Proceedings of the 64th Annual Meeting of the Association
  for Computational Linguistics (Volume 1: Long Papers)}, pages 7784--7812, San
  Diego, California, United States, July 2026. Association for Computational
  Linguistics.
\newblock \doi{10.18653/v1/2026.acl-long.354}.
\newblock URL \url{https://aclanthology.org/2026.acl-long.354/}.

\bibitem[Fleiss(1971)]{fleiss-1971}
Joseph~L. Fleiss.
\newblock Measuring nominal scale agreement among many raters.
\newblock \emph{Psychological Bulletin}, 76\penalty0 (5):\penalty0 378--382,
  1971.
\newblock \doi{10.1037/h0031619}.

\bibitem[{Google}(2025)]{google2025gemini25flash}
{Google}.
\newblock {Gemini 2.5 Flash}, 2025.
\newblock URL
  \url{https://ai.google.dev/gemini-api/docs/models/gemini-2.5-flash}.

\bibitem[{Google}(2026)]{google2026gemini31flashlite}
{Google}.
\newblock {Gemini 3.1 Flash-Lite}, 2026.
\newblock URL
  \url{https://ai.google.dev/gemini-api/docs/models/gemini-3.1-flash-lite}.

\bibitem[Gu et~al.(2024)Gu, Sun, Lian, Kang, Xu, and Fan]{gu-etal-2024-dingo}
Zihui Gu, Xingwu Sun, Fengzong Lian, Zhanhui Kang, Chengzhong Xu, and Ju~Fan.
\newblock {DINGO}: Towards diverse and fine-grained instruction-following
  evaluation.
\newblock In \emph{Proceedings of the AAAI Conference on Artificial
  Intelligence}, volume~38, pages 18108--18116, 2024.
\newblock \doi{10.1609/aaai.v38i16.29768}.
\newblock URL \url{https://ojs.aaai.org/index.php/AAAI/article/view/29768}.

\bibitem[Hossain et~al.(2020)Hossain, Kovatchev, Dutta, Kao, Wei, and
  Blanco]{hossain-etal-2020-analysis}
Md~Mosharaf Hossain, Venelin Kovatchev, Pranoy Dutta, Tiffany Kao, Elizabeth
  Wei, and Eduardo Blanco.
\newblock An analysis of natural language inference benchmarks through the lens
  of negation.
\newblock In \emph{Proceedings of the 2020 Conference on Empirical Methods in
  Natural Language Processing ({EMNLP})}, pages 9106--9118, Online, November
  2020. Association for Computational Linguistics.
\newblock \doi{10.18653/v1/2020.emnlp-main.732}.
\newblock URL \url{https://aclanthology.org/2020.emnlp-main.732/}.

\bibitem[Hua et~al.(2025)Hua, Tang, Gu, Gu, Wong, and Qin]{hua-etal-2025-flaw}
Andong Hua, Kenan Tang, Chenhe Gu, Jindong Gu, Eric Wong, and Yao Qin.
\newblock Flaw or artifact? rethinking prompt sensitivity in evaluating {LLM}s.
\newblock In \emph{Proceedings of the 2025 Conference on Empirical Methods in
  Natural Language Processing}, pages 19889--19899, Suzhou, China, November
  2025. Association for Computational Linguistics.
\newblock \doi{10.18653/v1/2025.emnlp-main.1006}.
\newblock URL \url{https://aclanthology.org/2025.emnlp-main.1006/}.

\bibitem[Jiang et~al.(2024)Jiang, Wang, Zeng, Zhong, Li, Mi, Shang, Jiang, Liu,
  and Wang]{jiang-etal-2024-followbench}
Yuxin Jiang, Yufei Wang, Xingshan Zeng, Wanjun Zhong, Liangyou Li, Fei Mi,
  Lifeng Shang, Xin Jiang, Qun Liu, and Wei Wang.
\newblock {F}ollow{B}ench: A multi-level fine-grained constraints following
  benchmark for large language models.
\newblock In \emph{Proceedings of the 62nd Annual Meeting of the Association
  for Computational Linguistics (Volume 1: Long Papers)}, pages 4667--4688,
  Bangkok, Thailand, August 2024. Association for Computational Linguistics.
\newblock \doi{10.18653/v1/2024.acl-long.257}.
\newblock URL \url{https://aclanthology.org/2024.acl-long.257/}.

\bibitem[Kassner and Sch{\"u}tze(2020)]{kassner-schutze-2020-negated}
Nora Kassner and Hinrich Sch{\"u}tze.
\newblock Negated and misprimed probes for pretrained language models: {B}irds
  can talk, but cannot fly.
\newblock In \emph{Proceedings of the 58th Annual Meeting of the Association
  for Computational Linguistics}, pages 7811--7818, Online, July 2020.
  Association for Computational Linguistics.
\newblock \doi{10.18653/v1/2020.acl-main.698}.
\newblock URL \url{https://aclanthology.org/2020.acl-main.698/}.

\bibitem[Kendall(1938)]{kendall-1938-new}
M.~G. Kendall.
\newblock A new measure of rank correlation.
\newblock \emph{Biometrika}, 30\penalty0 (1--2):\penalty0 81--93, 1938.
\newblock \doi{10.1093/biomet/30.1-2.81}.
\newblock URL \url{https://academic.oup.com/biomet/article/30/1-2/81/176907}.

\bibitem[Luccioni et~al.(2024)Luccioni, Jernite, and
  Strubell]{luccioni-etal-2024-power}
Alexandra~Sasha Luccioni, Yacine Jernite, and Emma Strubell.
\newblock Power hungry processing: Watts driving the cost of {AI} deployment?
\newblock In \emph{Proceedings of the 2024 ACM Conference on Fairness,
  Accountability, and Transparency}, pages 85--99, New York, NY, USA, 2024.
  Association for Computing Machinery.
\newblock \doi{10.1145/3630106.3658542}.
\newblock URL \url{https://dl.acm.org/doi/10.1145/3630106.3658542}.

\bibitem[{Mistral AI}(2026)]{mistral2026small4}
{Mistral AI}.
\newblock {Mistral Small 4} model card, 2026.
\newblock URL
  \url{https://docs.mistral.ai/models/model-cards/mistral-small-4-0-26-03}.

\bibitem[Mizrahi et~al.(2024)Mizrahi, Kaplan, Malkin, Dror, Shahaf, and
  Stanovsky]{mizrahi-etal-2024-state}
Moran Mizrahi, Guy Kaplan, Dan Malkin, Rotem Dror, Dafna Shahaf, and Gabriel
  Stanovsky.
\newblock State of what art? a call for multi-prompt {LLM} evaluation.
\newblock \emph{Transactions of the Association for Computational Linguistics},
  12:\penalty0 933--949, 2024.
\newblock \doi{10.1162/tacl_a_00681}.
\newblock URL \url{https://aclanthology.org/2024.tacl-1.52/}.

\bibitem[{Moonshot AI}(2026)]{moonshot2026k26}
{Moonshot AI}.
\newblock {Kimi K2.6} documentation, 2026.
\newblock URL \url{https://platform.kimi.ai/docs/guide/kimi-k2-6-quickstart}.

\bibitem[{OpenAI}(2024)]{openai2024gpt4omini}
{OpenAI}.
\newblock {GPT-4o mini} model, 2024.
\newblock URL \url{https://developers.openai.com/api/docs/models/gpt-4o-mini}.

\bibitem[{OpenAI}(2025)]{openai2025gpt41}
{OpenAI}.
\newblock {GPT-4.1 mini} model, 2025.
\newblock URL \url{https://developers.openai.com/api/docs/models/gpt-4.1-mini}.

\bibitem[{OpenAI}(2026)]{openai2026gpt55}
{OpenAI}.
\newblock {GPT-5.5} model, 2026.
\newblock URL \url{https://developers.openai.com/api/docs/models/gpt-5.5}.

\bibitem[Purpura et~al.(2026)Purpura, Wang, Badyal, Beaufrand, and
  Faulkner]{purpura-etal-2026-deconstructing}
Alberto Purpura, Li~Wang, Sahil Badyal, Gene Beaufrand, and Adam Faulkner.
\newblock Deconstructing instruction-following: A new benchmark for granular
  evaluation of large language model instruction compliance abilities.
\newblock In \emph{Proceedings of the 19th Conference of the European Chapter
  of the Association for Computational Linguistics (Volume 1: Long Papers)},
  pages 1334--1349, Rabat, Morocco, March 2026. Association for Computational
  Linguistics.
\newblock \doi{10.18653/v1/2026.eacl-long.62}.
\newblock URL \url{https://aclanthology.org/2026.eacl-long.62/}.

\bibitem[Pyatkin et~al.(2025)Pyatkin, Malik, Graf, Ivison, Huang, Dasigi,
  Lambert, and Hajishirzi]{pyatkin-etal-2025-ifbench}
Valentina Pyatkin, Saumya Malik, Victoria Graf, Hamish Ivison, Shengyi Huang,
  Pradeep Dasigi, Nathan Lambert, and Hanna Hajishirzi.
\newblock Generalizing verifiable instruction following.
\newblock In \emph{Advances in Neural Information Processing Systems},
  volume~38, 2025.
\newblock URL
  \url{https://proceedings.neurips.cc/paper_files/paper/2025/hash/46499a0622ecf568b72d17b61e45dbd5-Abstract-Datasets_and_Benchmarks_Track.html}.

\bibitem[Qin et~al.(2024)Qin, Song, Hu, Yao, Cho, Wang, Wu, Liu, Liu, and
  Yu]{qin-etal-2024-infobench}
Yiwei Qin, Kaiqiang Song, Yebowen Hu, Wenlin Yao, Sangwoo Cho, Xiaoyang Wang,
  Xuansheng Wu, Fei Liu, Pengfei Liu, and Dong Yu.
\newblock {I}n{F}o{B}ench: Evaluating instruction following ability in large
  language models.
\newblock In \emph{Findings of the Association for Computational Linguistics:
  ACL 2024}, pages 13025--13048, Bangkok, Thailand, August 2024. Association
  for Computational Linguistics.
\newblock \doi{10.18653/v1/2024.findings-acl.772}.
\newblock URL \url{https://aclanthology.org/2024.findings-acl.772/}.

\bibitem[{Qwen Team}(2026)]{qwen2026qwen36}
{Qwen Team}.
\newblock {Qwen3.6-35B-A3B} model card, 2026.
\newblock URL \url{https://huggingface.co/Qwen/Qwen3.6-35B-A3B}.

\bibitem[Ren et~al.(2026)Ren, Liang, Su, Sun, Lu, Zhang, and
  Wei]{ren-etal-2026-lexinstructeval}
Huimin Ren, Yan Liang, Baiqiao Su, Chaobo Sun, Hengtong Lu, Kaike Zhang, and
  Chen Wei.
\newblock {LexInstructEval}: Lexical instruction following evaluation for large
  language models.
\newblock In \emph{Proceedings of the AAAI Conference on Artificial
  Intelligence}, volume~40, pages 25117--25123, 2026.
\newblock \doi{10.1609/aaai.v40i30.39701}.
\newblock URL \url{https://ojs.aaai.org/index.php/AAAI/article/view/39701}.

\bibitem[Rezaei and Blanco(2024)]{rezaei-blanco-2024-paraphrasing}
MohammadHossein Rezaei and Eduardo Blanco.
\newblock Paraphrasing in affirmative terms improves negation understanding.
\newblock In \emph{Proceedings of the 62nd Annual Meeting of the Association
  for Computational Linguistics (Volume 2: Short Papers)}, pages 602--615,
  Bangkok, Thailand, August 2024. Association for Computational Linguistics.
\newblock \doi{10.18653/v1/2024.acl-short.55}.
\newblock URL \url{https://aclanthology.org/2024.acl-short.55/}.

\bibitem[Sclar et~al.(2024)Sclar, Choi, Tsvetkov, and
  Suhr]{sclar-etal-2024-quantifying}
Melanie Sclar, Yejin Choi, Yulia Tsvetkov, and Alane Suhr.
\newblock Quantifying language models' sensitivity to spurious features in
  prompt design or: How i learned to start worrying about prompt formatting.
\newblock In \emph{The Twelfth International Conference on Learning
  Representations}, 2024.
\newblock URL
  \url{https://proceedings.iclr.cc/paper_files/paper/2024/hash/6c0e99d736da621403018ca7b32b1a4d-Abstract-Conference.html}.

\bibitem[Shrout and Fleiss(1979)]{shrout-fleiss-1979}
Patrick~E. Shrout and Joseph~L. Fleiss.
\newblock Intraclass correlations: Uses in assessing rater reliability.
\newblock \emph{Psychological Bulletin}, 86\penalty0 (2):\penalty0 420--428,
  1979.
\newblock \doi{10.1037/0033-2909.86.2.420}.

\bibitem[Sun et~al.(2024)Sun, Shaib, and Wallace]{sun-etal-2024-evaluating}
Jiuding Sun, Chantal Shaib, and Byron~C. Wallace.
\newblock Evaluating the zero-shot robustness of instruction-tuned language
  models.
\newblock In \emph{The Twelfth International Conference on Learning
  Representations}, 2024.
\newblock URL
  \url{https://proceedings.iclr.cc/paper_files/paper/2024/hash/d3221cdb27e49d9c1cd35ad254feccfe-Abstract-Conference.html}.

\bibitem[Truong et~al.(2022)Truong, Otmakhova, Baldwin, Cohn, Lau, and
  Verspoor]{truong-etal-2022-another}
Hung~Thinh Truong, Yulia Otmakhova, Timothy Baldwin, Trevor Cohn, Jey~Han Lau,
  and Karin Verspoor.
\newblock Not another negation benchmark: The {N}a{N}-{NLI} test suite for
  sub-clausal negation.
\newblock In \emph{Proceedings of the 2nd Conference of the Asia-Pacific
  Chapter of the Association for Computational Linguistics and the 12th
  International Joint Conference on Natural Language Processing (Volume 1: Long
  Papers)}, pages 883--894, Online only, November 2022. Association for
  Computational Linguistics.
\newblock \doi{10.18653/v1/2022.aacl-main.65}.
\newblock URL \url{https://aclanthology.org/2022.aacl-main.65/}.

\bibitem[Truong et~al.(2023)Truong, Baldwin, Verspoor, and
  Cohn]{truong-etal-2023-language}
Thinh~Hung Truong, Timothy Baldwin, Karin Verspoor, and Trevor Cohn.
\newblock Language models are not naysayers: An analysis of language models on
  negation benchmarks.
\newblock In \emph{Proceedings of the 12th Joint Conference on Lexical and
  Computational Semantics ({*}{SEM} 2023)}, pages 101--114, Toronto, Canada,
  July 2023. Association for Computational Linguistics.
\newblock \doi{10.18653/v1/2023.starsem-1.10}.
\newblock URL \url{https://aclanthology.org/2023.starsem-1.10/}.

\bibitem[Webson and Pavlick(2022)]{webson-pavlick-2022-prompt}
Albert Webson and Ellie Pavlick.
\newblock Do prompt-based models really understand the meaning of their
  prompts?
\newblock In \emph{Proceedings of the 2022 Conference of the North American
  Chapter of the Association for Computational Linguistics: Human Language
  Technologies}, pages 2300--2344, Seattle, United States, July 2022.
  Association for Computational Linguistics.
\newblock \doi{10.18653/v1/2022.naacl-main.167}.
\newblock URL \url{https://aclanthology.org/2022.naacl-main.167/}.

\bibitem[Yang et~al.(2025)Yang, Liu, Guo, Xu, and
  Niu]{yang-etal-2025-enhancing}
Jiuding Yang, Hui Liu, Weidong Guo, Yu~Xu, and Di~Niu.
\newblock Enhancing and assessing instruction-following with fine-grained
  instruction variants.
\newblock In \emph{Proceedings of the 34th ACM International Conference on
  Information and Knowledge Management}, pages 3773--3782, New York, NY, USA,
  2025. Association for Computing Machinery.
\newblock \doi{10.1145/3746252.3761322}.
\newblock URL \url{https://dl.acm.org/doi/10.1145/3746252.3761322}.

\bibitem[Ye et~al.(2026)Ye, Huang, Chen, Fu, Yang, Yang, Wu, Wang, Zhou, Yang,
  Gui, Zhang, Shi, Fan, and Huang]{ye-etal-2026-muldimif}
Junjie Ye, Caishuang Huang, Zhuohan Chen, Wenjie Fu, Chenyuan Yang, Leyi Yang,
  Yilong Wu, Peng Wang, Meng Zhou, Xiaolong Yang, Tao Gui, Qi~Zhang, Zhongchao
  Shi, Jianping Fan, and Xuanjing Huang.
\newblock {M}ul{D}im{IF}: A multi-dimensional constraint framework for
  evaluating and improving instruction following in large language models.
\newblock In \emph{Findings of the Association for Computational Linguistics:
  ACL 2026}, pages 2078--2104, San Diego, California, United States, July 2026.
  Association for Computational Linguistics.
\newblock \doi{10.18653/v1/2026.findings-acl.99}.
\newblock URL \url{https://aclanthology.org/2026.findings-acl.99/}.

\bibitem[Yuan et~al.(2025)Yuan, Kulikov, Yu, Cho, Sukhbaatar, Weston, and
  Xu]{yuan-etal-2025-following}
Weizhe Yuan, Ilia Kulikov, Ping Yu, Kyunghyun Cho, Sainbayar Sukhbaatar,
  Jason~E Weston, and Jing Xu.
\newblock Following length constraints in instructions.
\newblock In \emph{Proceedings of the 2025 Conference on Empirical Methods in
  Natural Language Processing}, pages 24232--24243, Suzhou, China, November
  2025. Association for Computational Linguistics.
\newblock \doi{10.18653/v1/2025.emnlp-main.1233}.
\newblock URL \url{https://aclanthology.org/2025.emnlp-main.1233/}.

\bibitem[Zhou et~al.(2023)Zhou, Lu, Mishra, Brahma, Basu, Luan, Zhou, and
  Hou]{zhou-etal-2023-ifeval}
Jeffrey Zhou, Tianjian Lu, Swaroop Mishra, Siddhartha Brahma, Sujoy Basu,
  Yi~Luan, Denny Zhou, and Le~Hou.
\newblock Instruction-following evaluation for large language models, 2023.
\newblock URL \url{https://arxiv.org/abs/2311.07911}.

\bibitem[Zhuo et~al.(2024)Zhuo, Zhang, Fang, Duan, Lin, and
  Chen]{zhuo-etal-2024-prosa}
Jingming Zhuo, Songyang Zhang, Xinyu Fang, Haodong Duan, Dahua Lin, and Kai
  Chen.
\newblock {P}ro{SA}: Assessing and understanding the prompt sensitivity of
  {LLM}s.
\newblock In \emph{Findings of the Association for Computational Linguistics:
  EMNLP 2024}, pages 1950--1976, Miami, Florida, USA, November 2024.
  Association for Computational Linguistics.
\newblock \doi{10.18653/v1/2024.findings-emnlp.108}.
\newblock URL \url{https://aclanthology.org/2024.findings-emnlp.108/}.

\end{thebibliography}

\appendix

\section{Model and API Details}
\label{sec:model_details}

Table~\ref{tab:models} lists the evaluated models, exact API identifiers,
and access dates.

\begin{table*}[t]
\centering
\small
\begin{tabular}{llll}
\toprule
Provider & Paper name & API model ID & Accessed \\
\midrule
Anthropic & Claude Haiku 4.5 & \texttt{claude-haiku-4-5} & June 12, 2026 \\
Anthropic & Claude Sonnet 4.6 & \texttt{claude-sonnet-4-6} & June 12, 2026 \\
Anthropic & Claude Opus 4.8 & \texttt{claude-opus-4-8} & June 20, 2026 \\
OpenAI & GPT-4o mini & \texttt{gpt-4o-mini} & June 12, 2026 \\
OpenAI & GPT-4.1 mini & \texttt{gpt-4.1-mini} & June 12, 2026 \\
OpenAI & GPT-5.5 & \texttt{gpt-5.5-2026-04-23} & June 20, 2026 \\
Google & Gemini 2.5 Flash & \texttt{gemini-2.5-flash} & June 13, 2026 \\
Google & Gemini 3.1 Flash Lite & \texttt{gemini-3.1-flash-lite} & June 17, 2026 \\
DeepSeek & DeepSeek Chat & \texttt{deepseek-chat} & June 13, 2026 \\
DeepSeek & DeepSeek V4 Pro & \texttt{deepseek-v4-pro} & June 17, 2026 \\
Mistral & Mistral Small & \texttt{mistral-small-latest} & June 13, 2026 \\
Qwen & Qwen 3.6 35B A3B & \texttt{Qwen/Qwen3.6-35B-A3B} & June 17, 2026 \\
Kimi & Kimi K2.6 & \texttt{kimi-k2.6} & June 17, 2026 \\
\bottomrule
\end{tabular}
\caption{Evaluated models and API identifiers. Dynamic aliases are
reported exactly as called. The exact-count study uses all thirteen
models; the keyword and range extensions use the seven models available
when those runs were executed.}
\label{tab:models}
\end{table*}

\section{Paired Failure Examples}

Table~\ref{tab:matched_family_examples} gives matched examples in which
POS satisfies the strict count while another wording fails for the same
model, item, target, and repetition.

\begin{table*}[t]
\centering
\small
\setlength{\tabcolsep}{3.5pt}
\begin{tabular}{p{2.4cm}p{1.5cm}p{0.9cm}p{4.2cm}p{0.9cm}p{4.2cm}}
\toprule
Model & Form & Target & POS output & Count & Contrast output \\
\midrule
Claude Haiku 4.5
& AVOID1
& 5
& \emph{Jupiter is the largest planet.}
& 5
& \emph{Jupiter is the largest planet here.} (6, over) \\

DeepSeek Chat
& AVOID4
& 5
& \emph{Jupiter is the largest planet.}
& 5
& \emph{Jupiter} (1, under) \\

GPT-5.5
& EXCLUDE1
& 5
& \emph{Gold's chemical symbol is Au.}
& 5
& \emph{Au} (1, under) \\
\bottomrule
\end{tabular}
\caption{Matched avoidance-family examples in which POS is compliant but
an alternative wording fails for the same model, item, target, and
repetition. The examples include both over-target and under-target
failures.}
\label{tab:matched_family_examples}
\end{table*}

\section{Avoidance-Family Failure Examples}

Table~\ref{tab:avoid_examples} complements
Figure~\ref{fig:avoid_direction} with representative outputs. AVOID1 and
AVOID4 retain substantial over-target mass, while AVOID3, EXCLUDE1, and
EXCLUDE2 more often undershoot.

\begin{table*}[t]
\centering
\small
\begin{tabular}{p{2.1cm}p{1.6cm}p{1.2cm}p{1.5cm}p{7.0cm}}
\toprule
Model & Form & Target & Count & Output \\
\midrule
GPT-4o mini & AVOID1 & 5 & 6 &
\emph{The capital of France is Paris.} \\
GPT-4o mini & AVOID4 & 5 & 6 &
\emph{The capital of France is Paris.} \\
Claude Sonnet 4.6 & EXCLUDE1 & 5 & 4 &
\emph{William Shakespeare wrote it.} \\
Claude Sonnet 4.6 & EXCLUDE1 & 8 & 7 &
\emph{Vaccines train immunity, stopping transmission between people.} \\
GPT-4.1 mini & EXCLUDE2 & 5 & 4 &
\emph{William Shakespeare wrote Hamlet.} \\
\bottomrule
\end{tabular}
\caption{Representative failures from the avoidance-family panel under
strict entire-output whitespace counting.}
\label{tab:avoid_examples}
\end{table*}

\section{Surface-Behavior Audit}
\label{sec:surface_behavior}

\begin{figure}[t]
\centering
\includegraphics[width=\columnwidth]
{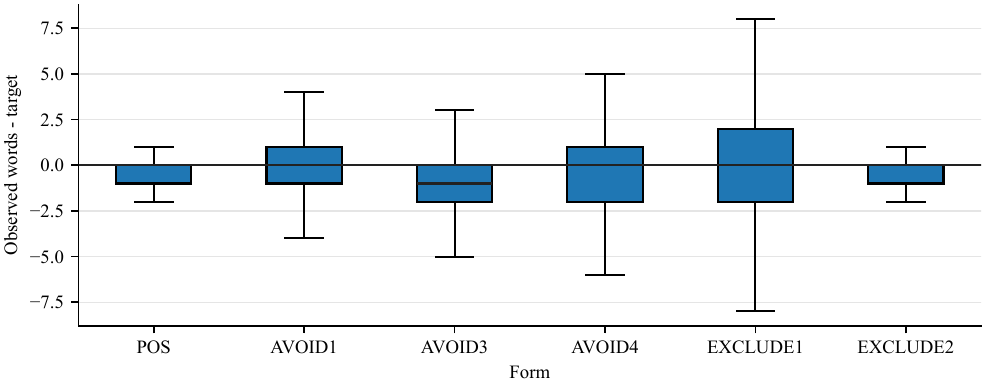}
\caption{Distribution of observed word count minus the requested target
in the avoidance-family panel for the eleven primary non-Gemini
models. AVOID3 and EXCLUDE2 are concentrated below the target, while
AVOID1, AVOID4, and especially EXCLUDE1 exhibit broader bidirectional
errors. Boxes show the interquartile range, whiskers extend to 1.5 times
the interquartile range, and outliers are not displayed.}
\label{fig:count_error_distribution}
\end{figure}

We examine visible output patterns in the avoidance-family panel
without using hidden states, token probabilities, or other model-internal
information. Across the eleven primary non-Gemini models, wording forms
produce distinct count-error distributions, as shown in
Figure~\ref{fig:count_error_distribution}. AVOID3 and EXCLUDE2 are
concentrated below the target, while AVOID1 and AVOID4 exhibit errors in
both directions. EXCLUDE1 has the broadest distribution and the highest
rate of outputs containing at most two words, 9.3\%, compared with 0.5\%
under POS.

Meta-discourse openings such as \emph{Here is}, \emph{Here's}, and
\emph{The answer is} are rare, occurring in fewer than 0.4\% of outputs
under every form. They therefore provide only limited descriptive context
and do not explain the main form-level compliance differences.

\section{Additional Figures}

\begin{figure}[t]
\centering
\includegraphics[width=\columnwidth]
{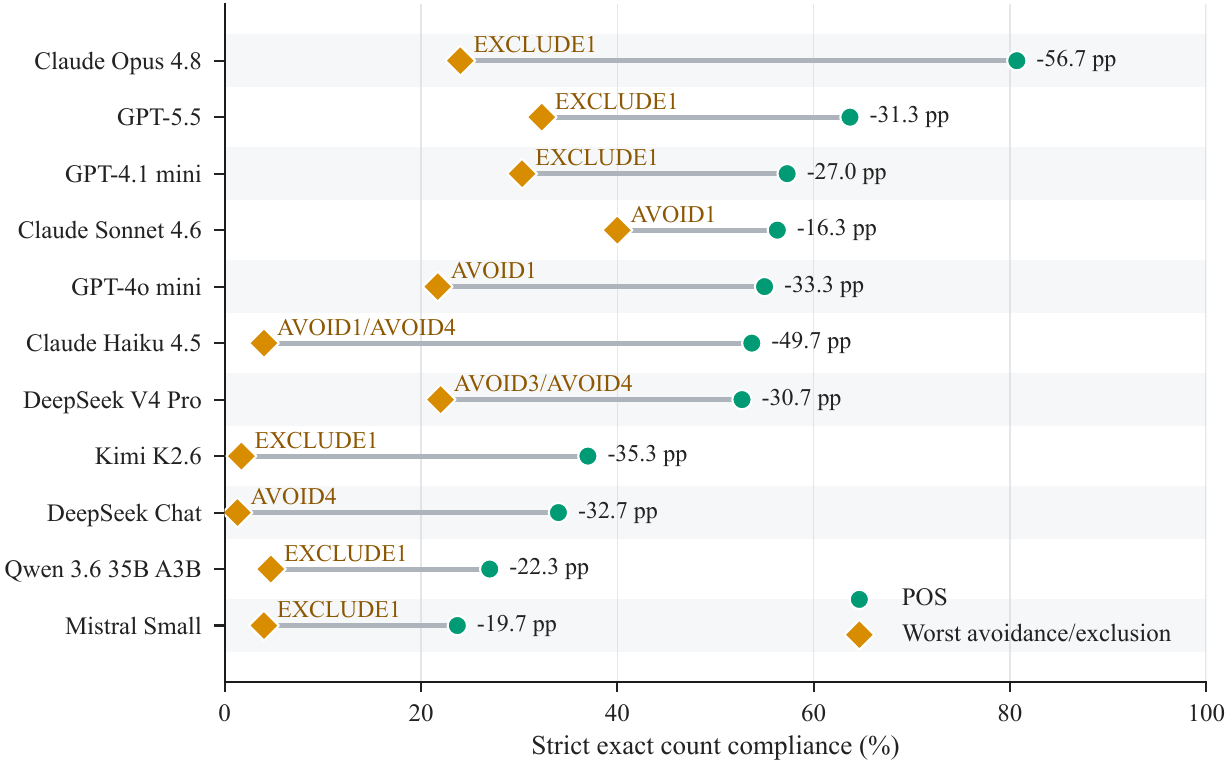}
\caption{POS-to-worst-form compliance gaps for the eleven primary
non-Gemini models in the avoidance-family panel. Circles show POS
compliance, diamonds show each model's weakest avoidance or exclusion
form, and annotations report the corresponding decrease.}
\label{fig:model_worst_gap_app}
\end{figure}

\begin{figure}[t]
\centering
\includegraphics[width=.95\linewidth]{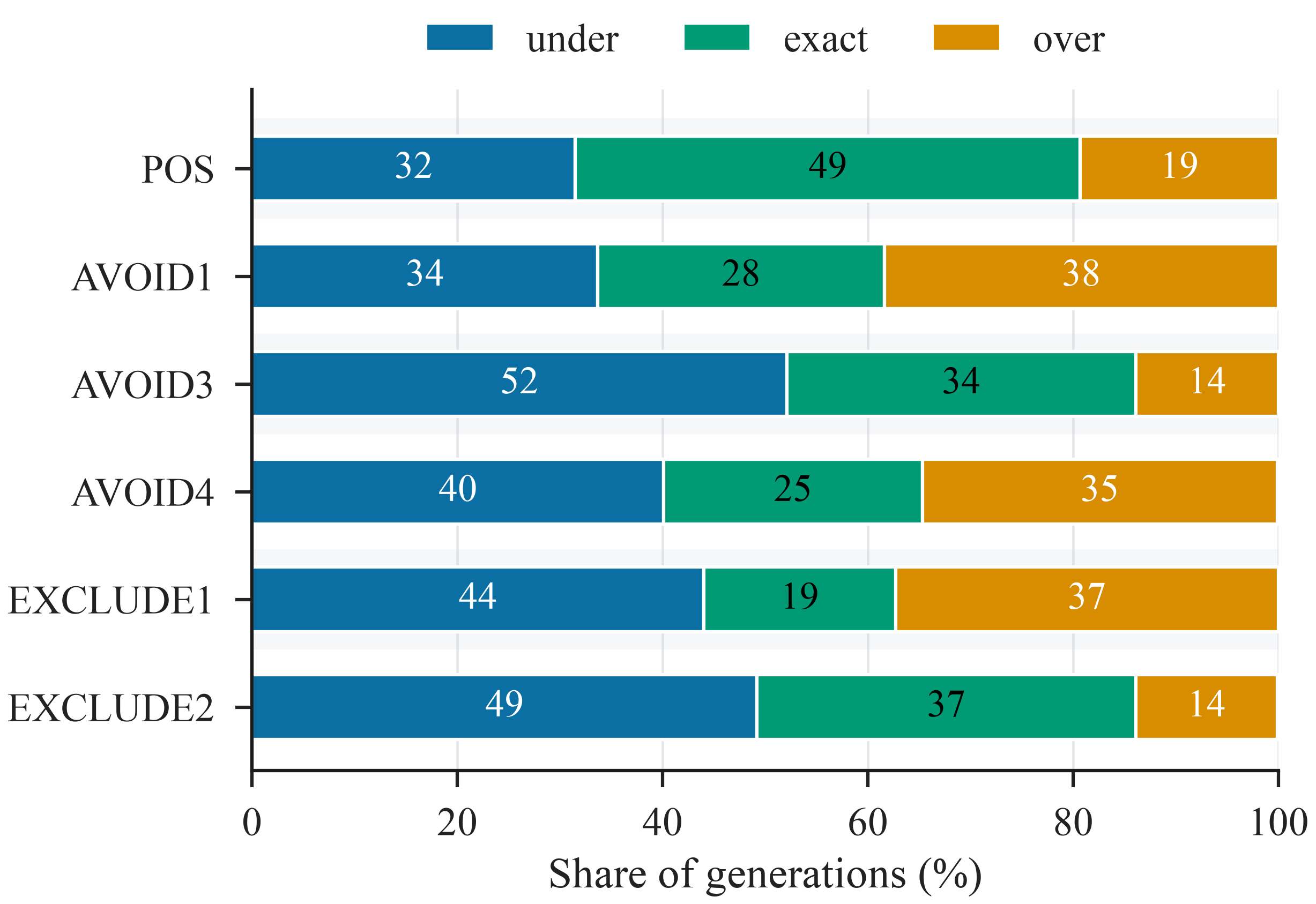}
\caption{Under-target, exact, and over-target rates in the
avoidance-family panel for the eleven non-Gemini models.}
\label{fig:avoid_direction}
\end{figure}

\begin{figure}[h]
\centering
\includegraphics[width=.95\linewidth]{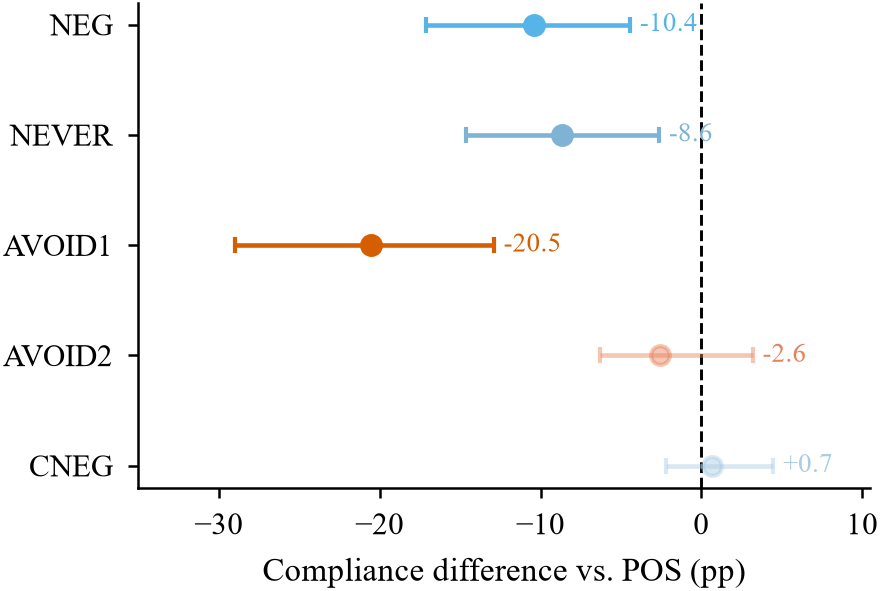}
\caption{Original-form compliance contrasts against POS for the eleven
non-Gemini models. Error bars are provider--model--item bootstrap 95\%
confidence intervals.}
\label{fig:pooled_original}
\end{figure}

\begin{figure}[t]
\centering
\includegraphics[width=.95\linewidth]{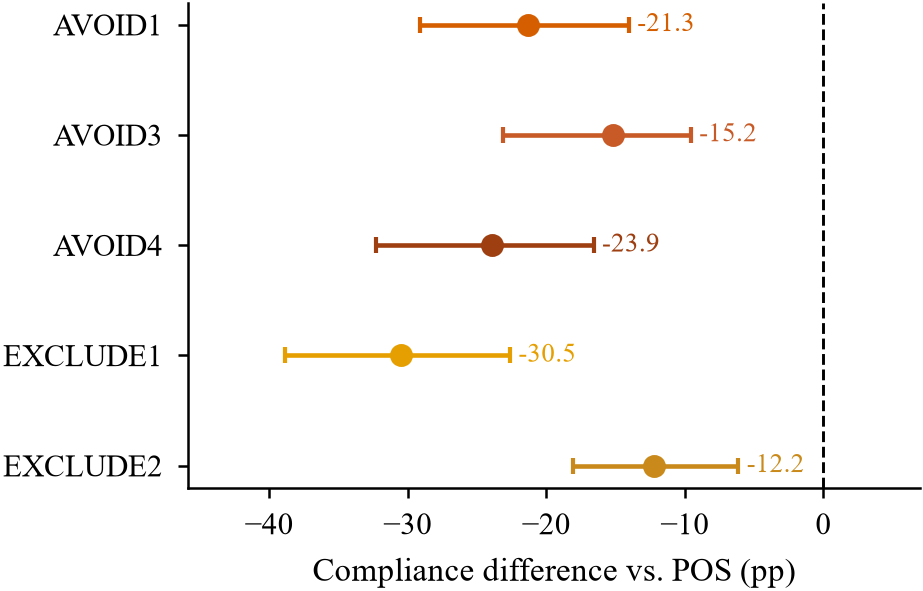}
\caption{Avoidance-family compliance contrasts against POS for the eleven
non-Gemini models.}
\label{fig:pooled_avoidance}
\end{figure}

Figures~\ref{fig:model_worst_gap_app}, \ref{fig:avoid_direction},
\ref{fig:pooled_original}, and \ref{fig:pooled_avoidance} visualize
model-specific gaps, failure directions, and pooled contrasts.

\section{Ranking Stability Metrics}
\label{sec:ranking_metrics}

Let
\begin{equation}
\bar{C}_{mf}
=
\frac{1}{IJ}
\sum_{i=1}^{I}
\sum_{j=1}^{J}
C_{ijmf}
\end{equation}
denote model \(m\)'s compliance under form \(f\). Relative to POS, the
strict pairwise reversal rate is
\begin{equation}
R_f
=
\frac{
\displaystyle
\sum_{a<b}
\mathbf{1}\!\left[
\left(
\bar{C}_{a,\mathrm{POS}}
-
\bar{C}_{b,\mathrm{POS}}
\right)
\left(
\bar{C}_{af}
-
\bar{C}_{bf}
\right)
<0
\right]
}{
\displaystyle
\sum_{a<b}
\mathbf{1}\!\left[
\bar{C}_{a,\mathrm{POS}}
\neq
\bar{C}_{b,\mathrm{POS}},
\,
\bar{C}_{af}
\neq
\bar{C}_{bf}
\right]
}.
\end{equation}
Pairs tied under either form are excluded. We report
Kendall's \(\tau_b\), top-three Jaccard overlap, and maximum absolute rank
displacement.

\section{Statistical Robustness}
\label{sec:multiple_comparison_appendix}

We assess multiplicity within each primary experimental family using
Holm correction. Raw two-sided bootstrap \(p\)-values are computed from
the same bootstrap draws used for the corresponding manuscript confidence
intervals. Table~\ref{tab:holm_results} reports the complete results.

\begin{table*}[t]
\centering
\small
\setlength{\tabcolsep}{4.5pt}
\renewcommand{\arraystretch}{1.05}
\begin{tabular}{llrrrr}
\toprule
Family & Contrast & Diff. & 95\% CI & Raw \(p\) & Holm \(p\) \\
\midrule
Original exact-count
& NEG--POS
& -10.4 & [-17.1, -4.4] & <.001 & <.001 \\

& NEVER--POS
& -8.6 & [-14.6, -2.6] & .006 & .018 \\

& AVOID1--POS
& -20.5 & [-29.0, -12.9] & <.001 & <.001 \\

& AVOID2--POS
& -2.6 & [-6.3, 3.2] & .357 & .714 \\

& CNEG--POS
& +0.7 & [-2.2, 4.5] & .689 & .714 \\
\midrule
Avoidance family
& AVOID1--POS
& -21.3 & [-29.1, -14.0] & <.001 & <.001 \\

& AVOID3--POS
& -15.2 & [-23.2, -9.6] & <.001 & <.001 \\

& AVOID4--POS
& -23.9 & [-32.3, -16.6] & <.001 & <.001 \\

& EXCLUDE1--POS
& -30.5 & [-38.9, -22.6] & <.001 & <.001 \\

& EXCLUDE2--POS
& -12.2 & [-18.1, -6.1] & <.001 & <.001 \\
\midrule
Keyword primary
& NEG--POS
& -22.1 & [-25.8, -18.2] & <.001 & <.001 \\

& AVOID--POS
& -33.2 & [-37.3, -29.1] & <.001 & <.001 \\
\midrule
Range primary
& NEG--POS
& -3.6 & [-10.6, 1.2] & .237 & .237 \\

& AVOID--POS
& -8.8 & [-16.3, -3.2] & <.001 & <.001 \\
\bottomrule
\end{tabular}
\caption{Multiplicity analysis for the primary wording contrasts.
Differences are percentage points relative to POS. Raw two-sided
bootstrap \(p\)-values and 95\% confidence intervals are computed from
the same bootstrap draws used in the primary analysis. Holm correction
is applied separately within each experimental family.}
\label{tab:holm_results}
\end{table*}

\section{\wise{} Audit Summary}
\label{sec:audit_appendix}

Table~\ref{tab:audit_summary} reports the complete audit summaries for
the original exact-count panel, avoidance-family panel, constructional
controls, keyword constraint, inclusive length range, and JSON probe.

\begin{table}[t]
\centering
\small
\setlength{\tabcolsep}{3.5pt}
\renewcommand{\arraystretch}{1.05}
\begin{tabular}{lrrrr}
\toprule
Panel & POS & \(C_{\mathrm{mean}}\) &
\(C_{\mathrm{worst}}\) & \(G_{\mathrm{wording}}\) \\
\midrule
Original exact count & 49.5 & 42.5 & 28.9 & 21.2 \\
Avoidance family     & 49.2 & 32.0 & 18.7 & 30.5 \\
Constructional controls & 54.9 & 43.4 & 33.8 & 21.1 \\
Keyword exactly once & 77.4 & 59.0 & 44.3 & 33.2 \\
Inclusive 8--12 range& 89.4 & 85.3 & 80.6 & 8.8 \\
JSON structure       & 13.1 & 16.6 & 13.1 & 7.6 \\
\bottomrule
\end{tabular}
\caption{\wise{} audit summaries across experimental panels.
Values are percentages over each panel's primary model set. Gaps are
computed from unrounded rates. For JSON, POS is the original positive
form even though it is not the best-performing form.}
\label{tab:audit_summary}
\end{table}

\section{Gemini Saved-Output Diagnostics}
\label{sec:gemini_diagnostics}

The two Gemini models require separate interpretation. Their saved outputs
are very short across forms, and exact compliance is at or near zero for
targets 8, 10, and 12. We define a saved-output floor regime as exact
compliance of at most 1\% at every target above the minimum and a change
of less than one mean observed word as the requested target increases from
5 to 12. Both Gemini models satisfy this criterion in both experiments.
Table~\ref{tab:gemini_floor} gives the Gemini 2.5 target-level results;
Gemini 3.1 shows the same qualitative pattern, with exact compliance of
21.6\%, 0.9\%, 0.7\%, and 0.0\% at targets 5, 8, 10, and 12 in the
original experiment and mean output length between 2.36 and 2.88 words.

A post hoc audit finds no API-error rows, and saved output-token counts
are far below the 120-token limit, at most 9 tokens for Gemini 2.5 and 15
for Gemini 3.1, which provides no evidence of length-limit truncation.
Gemini 2.5 has no empty saved outputs; Gemini 3.1 has empty-text rates of
9.7\% and 10.7\% in the original and avoidance-family runs. The adapter excludes
usage metadata from evaluated text. The runs did not retain raw response
bodies, \texttt{finishReason}, or per-part thought flags, so
provider-specific response-part extraction cannot be fully excluded as a
contributor to short or empty saved outputs. We therefore exclude both
Gemini models from the main wording-specific contrasts.

\begin{table}[t]
\centering
\small
\setlength{\tabcolsep}{3.5pt}
\begin{tabular}{lrrrr}
\toprule
Experiment & Target & Exact & Mean words & Mean error \\
\midrule
Original & 5  & 30.4 & 3.66 & -1.34 \\
Original & 8  & 0.0  & 2.85 & -5.15 \\
Original & 10 & 0.0  & 2.89 & -7.11 \\
Original & 12 & 0.0  & 2.93 & -9.07 \\
Family   & 5  & 30.9 & 3.56 & -1.44 \\
Family   & 8  & 0.0  & 2.76 & -5.24 \\
Family   & 10 & 0.0  & 2.91 & -7.09 \\
Family   & 12 & 0.0  & 2.90 & -9.10 \\
\bottomrule
\end{tabular}
\caption{Gemini 2.5 Flash saved-output floor analysis. Exact is strict
compliance in percent; mean error is observed word count minus requested
target.}
\label{tab:gemini_floor}
\end{table}

\section{Cross-Constraint Pooled Results}
\label{sec:cross_constraint_robustness}

Tables~\ref{tab:keyword_primary} and
\ref{tab:length_range_primary} report the primary pooled analyses for the
six non-Gemini models shared by the keyword and range extensions.
Tables~\ref{tab:keyword_all_models} and
\ref{tab:length_range} report the corresponding seven-model analyses,
including Gemini 2.5 Flash.

\begin{table}[t]
\centering
\small
\setlength{\tabcolsep}{4.2pt}
\renewcommand{\arraystretch}{1.05}
\begin{tabular}{lrrr}
\toprule
Form & Compliance & Diff. & 95\% CI \\
\midrule
POS   & 77.4 & $-$   & $-$             \\
NEG   & 55.3 & -22.1 & [-25.8, -18.2] \\
AVOID & 44.3 & -33.2 & [-37.3, -29.1] \\
\bottomrule
\end{tabular}
\caption{Primary keyword-exactly-once results for the six shared
non-Gemini models. Values are percentages; differences and item-cluster
bootstrap 95\% confidence intervals are relative to POS.}
\label{tab:keyword_primary}
\end{table}

\begin{table}[t]
\centering
\small
\setlength{\tabcolsep}{4.2pt}
\renewcommand{\arraystretch}{1.05}
\begin{tabular}{lrrr}
\toprule
Form & Compliance & Diff. & 95\% CI \\
\midrule
POS   & 89.4 & $-$  & $-$            \\
NEG   & 85.8 & -3.6 & [-10.6, 1.2]  \\
AVOID & 80.6 & -8.8 & [-16.3, -3.2] \\
\bottomrule
\end{tabular}
\caption{Primary inclusive 8--12 word range results for the six shared
non-Gemini models. Values are percentages; differences and model--item
bootstrap 95\% confidence intervals are relative to POS.}
\label{tab:length_range_primary}
\end{table}

\begin{table}[t]
\centering
\small
\setlength{\tabcolsep}{2.6pt}
\renewcommand{\arraystretch}{1.05}
\begin{tabular}{@{}lrrrrr@{}}
\toprule
Form & Comp. & Diff. & 95\% CI & Zero & Multi. \\
\midrule
POS   & 69.7 & $-$   & $-$             & 22.8 & 7.5 \\
NEG   & 49.2 & -20.5 & [-24.0, -16.7] & 47.5 & 3.2 \\
AVOID & 38.7 & -31.1 & [-34.8, -27.3] & 58.9 & 2.5 \\
\bottomrule
\end{tabular}
\caption{Keyword-exactly-once robustness across all seven models,
including Gemini 2.5 Flash. Values are percentages. Diff. and confidence
intervals are relative to POS; Zero and Multi. denote zero and multiple
keyword occurrences.}
\label{tab:keyword_all_models}
\end{table}

\begin{table}[t]
\centering
\small
\setlength{\tabcolsep}{3.6pt}
\renewcommand{\arraystretch}{1.05}
\begin{tabular}{lrrrrr}
\toprule
Form & Comp. & Diff. & 95\% CI & Below & Above \\
\midrule
POS   & 76.6 & $-$  & $-$            & 17.2 & 6.2 \\
NEG   & 73.6 & -3.0 & [-9.4, 1.0]    & 17.1 & 9.3 \\
AVOID & 69.1 & -7.5 & [-14.3, -2.2]  & 22.0 & 8.8 \\
\bottomrule
\end{tabular}
\caption{Inclusive 8--12 word range robustness across all seven models,
including Gemini 2.5 Flash. Values are percentages; intervals use the
model--item bootstrap.}
\label{tab:length_range}
\end{table}

\section{Constructional-Control Model-Level Results}
\label{sec:positive_control_appendix}

Table~\ref{tab:positive_control_models} reports the four-form
exact-count constructional-control panel by model. The primary analysis
uses the nine retained models shown here and excludes Qwen and Kimi
because their endpoint/configuration could not be matched to the original
collection. Including all eleven models preserves the pooled ordering.

\begin{table}[t]
\centering
\small
\resizebox{\columnwidth}{!}{
\begin{tabular}{lrrrr}
\toprule
Model & POS & POS\_LONG & POS\_END & AVOID1 \\
\midrule
Claude Haiku 4.5  & 49.3 & 34.7 & 21.7 & 5.0 \\
Claude Opus 4.8   & 80.7 & 64.3 & 35.7 & 83.0 \\
Claude Sonnet 4.6 & 59.3 & 63.0 & 60.0 & 44.7 \\
DeepSeek Chat     & 49.3 & 34.3 & 16.0 & 26.3 \\
DeepSeek V4 Pro   & 59.3 & 58.7 & 46.0 & 28.7 \\
GPT-4.1 mini      & 53.7 & 53.0 & 47.7 & 39.3 \\
GPT-4o mini       & 58.0 & 48.3 & 31.3 & 24.0 \\
GPT-5.5           & 58.3 & 55.0 & 56.7 & 42.0 \\
Mistral Small     & 26.3 & 22.7 & 15.3 & 11.3 \\
\bottomrule
\end{tabular}
}
\caption{Per-model compliance in the constructional-control panel. Values
are percentages.}
\label{tab:positive_control_models}
\end{table}

\section{JSON-Structure Model-Level Results}
\label{sec:json_appendix}

Table~\ref{tab:json_model_form} reports the strict JSON-structure probe
by model and form. The main failure mode is unparseable complete visible
output rather than wrong key sets.

\begin{table}[t]
\centering
\small
\begin{tabular}{lrrr}
\toprule
Model & POS & NEG & AVOID \\
\midrule
Claude Haiku 4.5  & 0.0 & 0.0 & 0.0 \\
Claude Sonnet 4.6 & 0.0 & 0.0 & 0.0 \\
DeepSeek Chat     & 70.7 & 62.0 & 62.3 \\
GPT-4.1 mini      & 0.7 & 6.0 & 5.3 \\
GPT-4o mini       & 7.0 & 28.7 & 55.0 \\
Mistral Small     & 0.0 & 0.0 & 1.0 \\
\bottomrule
\end{tabular}
\caption{Strict JSON-structure compliance by model and wording form.
Values are percentages.}
\label{tab:json_model_form}
\end{table}

\section{Length-Range Model-Level Results}
\label{sec:length_range_appendix}

Table~\ref{tab:length_range_models} reports the complete model-by-form
results. Claude Sonnet is near ceiling and Gemini 2.5 Flash is at floor,
while the remaining models show varying gaps. The experiment contains
6,300 unique generations, zero API errors, zero missing rows, and zero
duplicate keys.

\begin{table}[t]
\centering
\small
\setlength{\tabcolsep}{3.2pt}
\renewcommand{\arraystretch}{1.05}
\begin{tabular}{lrrrr}
\toprule
Model & POS & NEG & AVOID & Gap \\
\midrule
Claude Haiku 4.5  & 73.7 & 54.3 & 49.3 & 24.3 \\
Claude Sonnet 4.6 & 100.0 & 100.0 & 98.7 & 1.3 \\
GPT-4o mini       & 97.0 & 95.3 & 88.3 & 8.7 \\
GPT-4.1 mini      & 97.0 & 97.3 & 93.3 & 4.0 \\
Gemini 2.5 Flash  & 0.0 & 0.0 & 0.7 & 0.7 \\
DeepSeek Chat     & 92.0 & 90.3 & 89.0 & 3.0 \\
Mistral Small     & 76.7 & 77.7 & 64.7 & 13.0 \\
\bottomrule
\end{tabular}
\caption{Strict 8--12 word range compliance by model and wording form.
Gap is the best--worst difference in percentage points.}
\label{tab:length_range_models}
\end{table}

\section{Keyword Model-Level Results}

Table~\ref{tab:keyword_model_form} reports the complete model-by-form
keyword results.

\begin{table}[t]
\centering
\small
\begin{tabular}{lrrr}
\toprule
Model & POS & NEG & AVOID \\
\midrule
Claude Haiku 4.5  & 78.7 & 77.7 & 77.7 \\
Claude Sonnet 4.6 & 88.0 & 88.3 & 91.0 \\
GPT-4o mini       & 70.3 & 12.3 & 9.3 \\
GPT-4.1 mini      & 91.7 & 79.7 & 38.3 \\
Gemini 2.5 Flash  & 23.3 & 12.7 & 5.0 \\
DeepSeek Chat     & 73.7 & 44.3 & 41.0 \\
Mistral Small     & 62.3 & 29.7 & 8.3 \\
\bottomrule
\end{tabular}
\caption{Keyword-exactly-once strict compliance by model and wording form.
Values are percentages. These are the seven models included when the
keyword experiment was run.}
\label{tab:keyword_model_form}
\end{table}

\section{Exact-Count Model-Level Results}
\label{sec:exact_count_model_results}

Tables~\ref{tab:appendix_original_model_form} and
\ref{tab:appendix_family_model_form} report complete model-by-form
compliance values. These tables include Gemini; the main text separates
the Gemini models because they satisfy the floor-effect criterion in
Section~\ref{sec:gemini_diagnostics}.

\begin{table}[t]
\centering
\small
\resizebox{\columnwidth}{!}{%
\begin{tabular}{lrrrrrr}
\toprule
Model & POS & NEG & NEVER & AVOID1 & AVOID2 & CNEG \\
\midrule
Claude Haiku 4.5  & 54.7 & 38.3 & 38.7 & 4.7  & 54.0 & 52.7 \\
Claude Sonnet 4.6 & 58.3 & 60.0 & 67.0 & 41.7 & 46.7 & 60.7 \\
Claude Opus 4.8   & 82.3 & 78.3 & 79.3 & 79.0 & 81.3 & 80.3 \\
GPT-4o mini       & 53.7 & 38.3 & 38.3 & 21.0 & 48.0 & 54.7 \\
GPT-4.1 mini      & 55.3 & 39.0 & 39.3 & 44.0 & 54.3 & 60.0 \\
GPT-5.5           & 62.7 & 67.0 & 64.7 & 43.0 & 52.7 & 56.7 \\
Gemini 2.5 Flash  & 9.7  & 5.7  & 8.7  & 9.3  & 6.7  & 5.7 \\
Gemini 3.1 Flash Lite & 6.7 & 6.3 & 4.0 & 5.3 & 6.3 & 6.0 \\
DeepSeek Chat     & 36.3 & 28.7 & 31.7 & 13.3 & 39.3 & 39.3 \\
DeepSeek V4 Pro   & 55.0 & 29.7 & 35.3 & 38.0 & 48.7 & 51.7 \\
Mistral Small     & 24.7 & 7.7  & 9.3  & 10.7 & 19.7 & 24.7 \\
Qwen 3.6 35B A3B & 26.7 & 25.0 & 24.3 & 16.7 & 26.7 & 27.0 \\
Kimi K2.6         & 34.3 & 17.3 & 21.0 & 6.0 & 44.3 & 43.7 \\
\bottomrule
\end{tabular}%
}
\caption{Original six-form strict compliance by model and form. Values are
percentages.}
\label{tab:appendix_original_model_form}
\end{table}

\begin{table}[t]
\centering
\small
\resizebox{\columnwidth}{!}{%
\begin{tabular}{lrrrrrr}
\toprule
Model & POS & AVOID1 & AVOID3 & AVOID4 & EXCLUDE1 & EXCLUDE2 \\
\midrule
Claude Haiku 4.5  & 53.7 & 4.0  & 36.3 & 4.0  & 7.7  & 36.0 \\
Claude Sonnet 4.6 & 56.3 & 40.0 & 57.7 & 50.3 & 41.7 & 54.3 \\
Claude Opus 4.8   & 80.7 & 78.3 & 67.7 & 59.7 & 24.0 & 67.0 \\
GPT-4o mini       & 55.0 & 21.7 & 40.0 & 27.3 & 23.3 & 27.7 \\
GPT-4.1 mini      & 57.3 & 42.7 & 44.3 & 41.0 & 30.3 & 37.3 \\
GPT-5.5           & 63.7 & 42.7 & 57.7 & 54.3 & 32.3 & 62.3 \\
Gemini 2.5 Flash  & 8.7  & 8.7  & 6.7  & 8.3  & 5.0  & 9.0 \\
Gemini 3.1 Flash Lite & 7.3 & 4.3 & 6.0 & 7.3 & 5.3 & 7.3 \\
DeepSeek Chat     & 34.0 & 12.0 & 23.3 & 1.3  & 4.0  & 33.3 \\
DeepSeek V4 Pro   & 52.7 & 39.0 & 22.0 & 22.0 & 31.7 & 32.7 \\
Mistral Small     & 23.7 & 9.3  & 6.3  & 6.0  & 4.0  & 14.3 \\
Qwen 3.6 35B A3B & 27.0 & 12.3 & 10.0 & 6.3 & 4.7 & 21.0 \\
Kimi K2.6         & 37.0 & 4.7 & 8.3 & 5.3 & 1.7 & 20.3 \\
\bottomrule
\end{tabular}%
}
\caption{Avoidance-family strict compliance by model and form. Values are
percentages.}
\label{tab:appendix_family_model_form}
\end{table}

\section{Bootstrap Sensitivity Analyses}
\label{sec:bootstrap_sensitivity}

The main text uses provider--model--item hierarchical bootstrap intervals.
Because there are only six providers in the non-Gemini analysis and the
hierarchy is unbalanced, we also compute two sensitivity analyses on the
non-Gemini set: a model--item bootstrap, which treats model as the highest
resampling level, and an equal-model item bootstrap, which bootstraps
items within each model and averages model-level contrasts with equal
weight. Table~\ref{tab:bootstrap_sensitivity} shows that the main
conclusions hold for the key contrasts.

\begin{table}[t]
\centering
\small
\resizebox{\columnwidth}{!}{%
\begin{tabular}{llllr}
\toprule
Experiment & Contrast & Method & 95\% CI & Diff. \\
\midrule
Original & AVOID1 $-$ POS & Provider--model--item & [-29.0, -12.9] & -20.5 \\
Original & AVOID1 $-$ POS & Model--item & [-28.7, -13.6] & -20.5 \\
Original & AVOID1 $-$ POS & Equal-model item & [-23.1, -17.8] & -20.5 \\
Original & NEG $-$ POS & Provider--model--item & [-17.1, -4.4] & -10.4 \\
Original & NEG $-$ POS & Model--item & [-16.0, -4.6] & -10.4 \\
Original & NEG $-$ POS & Equal-model item & [-12.9, -7.9] & -10.4 \\
Family & AVOID4 $-$ POS & Provider--model--item & [-32.3, -16.6] & -23.9 \\
Family & AVOID4 $-$ POS & Model--item & [-31.5, -17.1] & -23.9 \\
Family & AVOID4 $-$ POS & Equal-model item & [-26.7, -21.3] & -23.9 \\
Family & EXCLUDE1 $-$ POS & Provider--model--item & [-38.9, -22.6] & -30.5 \\
Family & EXCLUDE1 $-$ POS & Model--item & [-38.4, -23.5] & -30.5 \\
Family & EXCLUDE1 $-$ POS & Equal-model item & [-33.2, -27.7] & -30.5 \\
\bottomrule
\end{tabular}%
}
\caption{Bootstrap sensitivity analyses for key non-Gemini contrasts.
Differences and confidence intervals are percentage points.}
\label{tab:bootstrap_sensitivity}
\end{table}

\section{Human Wording Validation}
\label{sec:human_validation}

We conducted a blinded wording-validation study to examine whether the
six exact-count forms are perceived as expressing the same operational
requirement. The validation set contains 24 randomized items, crossing
six forms with target counts of 5, 8, 10, and 12 words. Annotators saw
neutral item identifiers, target counts, and instructions without form
labels.

For each item, annotators judged whether the instruction required exactly
\(N\) words and provided ratings of clarity, naturalness, ambiguity, and
complexity on five-point scales. Five annotators completed all 24 items,
yielding 120 judgments for each measure.

\subsection{Annotator-Level Exact-Count Judgments}

\begin{table}[t]
\centering
\small
\setlength{\tabcolsep}{3.4pt}
\renewcommand{\arraystretch}{1.07}
\begin{tabular}{@{}lrrrrrr@{}}
\toprule
Form & A1 & A2 & A3 & A4 & A5 & Exact-\(N\) \\
\midrule
POS
& 1.00 & 1.00 & 1.00 & 1.00 & 1.00 & 100 \\

AVOID1
& 1.00 & 0.75 & 1.00 & 0.75 & 1.00 & 90 \\

AVOID3
& 1.00 & 1.00 & 0.75 & 1.00 & 1.00 & 95 \\

AVOID4
& 1.00 & 1.00 & 0.75 & 0.75 & 1.00 & 90 \\

EXCLUDE1
& 1.00 & 0.75 & 1.00 & 0.75 & 1.00 & 90 \\

EXCLUDE2
& 1.00 & 1.00 & 1.00 & 1.00 & 1.00 & 100 \\
\bottomrule
\end{tabular}
\caption{Annotator-level exact-count judgment rates by wording form. Each
annotator entry is the proportion of exact-\(N\) judgments across target
counts of 5, 8, 10, and 12 words. The final column aggregates all twenty
judgments for each form.}
\label{tab:human_individual}
\end{table}

Table~\ref{tab:human_individual} shows that the exact-count interpretation
is broadly shared. POS and EXCLUDE2 receive exact-\(N\) judgments for all
twenty annotator and target combinations. The four remaining forms receive
exact-\(N\) judgments in 90\% to 95\% of combinations. Variation is
therefore limited to a small number of target-specific judgments rather
than a general rejection of the exact-count interpretation.

\subsection{Inter-Annotator Agreement}

We summarize agreement using Fleiss' \(\kappa\) for the binary
exact-\(N\) judgment \citep{fleiss-1971} and ICC(2,\(k\)) for the
continuous ratings \citep{shrout-fleiss-1979}.

\begin{table}[t]
\centering
\small
\setlength{\tabcolsep}{4.6pt}
\renewcommand{\arraystretch}{1.07}
\begin{tabular}{@{}llrr@{}}
\toprule
Measure & Statistic & Estimate & 95\% CI \\
\midrule
Exact-\(N\) judgment
& Fleiss' \(\kappa\)
& 0.54
& [0.34, 0.72] \\

Clarity
& ICC(2,\(k\))
& 0.79
& [0.62, 0.90] \\

Naturalness
& ICC(2,\(k\))
& 0.75
& [0.56, 0.88] \\

Ambiguity
& ICC(2,\(k\))
& 0.68
& [0.45, 0.84] \\

Complexity
& ICC(2,\(k\))
& 0.64
& [0.40, 0.81] \\
\bottomrule
\end{tabular}
\caption{Inter-annotator agreement across all 24 wording and target
combinations. Fleiss' \(\kappa\) is used for the binary exact-\(N\)
judgment. Continuous ratings use ICC(2,\(k\)), a two-way random-effects,
absolute-agreement, average-measures specification. Confidence intervals
are obtained from 2000 bootstrap replicates that resample the 24 items
while retaining all five annotators for each sampled item.}
\label{tab:human_agreement}
\end{table}

Table~\ref{tab:human_agreement} reports moderate agreement for the binary
exact-\(N\) judgment, with Fleiss' \(\kappa=0.54\) and a bootstrap
95\% interval of [0.34, 0.72].
Agreement is higher for clarity and naturalness, with ICC values of
0.79 and 0.75, and somewhat lower for ambiguity and complexity, with
ICC values of 0.68 and 0.64. These statistics support using the ratings
as descriptive evidence while preserving uncertainty about fine-grained
differences among forms.

The central behavioral comparison does not depend on fine-grained rating
agreement. POS and EXCLUDE2 receive unanimous exact-\(N\) judgments, yet
EXCLUDE2 lowers pooled model compliance by 12.2 percentage points relative
to POS. AVOID1, AVOID4, and EXCLUDE1 also receive 90\% exact-\(N\)
judgments while producing considerably larger compliance reductions.
Human recognition of the target and model compliance are therefore
related but distinct properties.

\section{Reproducibility Details}
\label{sec:reproducibility}

All experiments use the same 100-task manifest, which records the item ID,
base task, and target count. Separate form manifests record the wording
templates, and the released raw generation files include the instantiated
instruction and complete prompt for each generation. The original
exact-count study and its avoidance-family panel pair each task with six
wording forms and collect three generations per model and form. The
constructional-control panel uses four forms and three generations for
each retained model. The keyword and range extensions use
three forms and three generations for each of seven models, and the JSON
probe uses three forms with a strict full-output parser.

The exact-count and range checkers count whitespace-delimited tokens using
\texttt{len((output or "").strip().split())}. Any Unicode whitespace
recognized by Python separates words, while multiple spaces and newlines
collapse under \texttt{split()}. Punctuation-attached strings,
contractions, hyphenated words, Markdown markers, headings, and wrappers
count as visible tokens when separated by whitespace; empty or
whitespace-only output has count zero.

The released \wise{} package includes raw generations, strict verifier
results, word or keyword counts, compliance indicators, directional
failure flags, bootstrap summaries, model-level profiles, paired examples,
constructional-control and JSON-control summaries, and audit summaries. The
\wise{} audit script combines the constraint panels while preserving
their constraint-specific metrics.
The human-validation release contains verified form-level summaries,
annotator-level exact-\(N\) rates, and agreement statistics, together with
release scripts that reproduce the reported tables.

\clearpage


\section*{Supplementary material}

# Qualitative Examples of Wording-Sensitive Failures

Matched examples from the released generation outputs.

### Longer Affirmative Form Changes the Count

**Model:** GPT-4o mini **Target:** 5 words

**Task:** Name the process plants use to make food.

| POS output (count = 5) | POS_LONG output (count = 6, over) |
|---|---|
| Photosynthesis converts sunlight into food. | Photosynthesis is how plants make food. |

Same polarity and same verifier; the longer affirmative realization fails while POS succeeds.

### Late Target Placement in an Affirmative Form

**Model:** GPT-4.1 mini **Target:** 5 words

**Task:** What is H2O commonly called?

| POS output (count = 5) | POS_END output (count = 4, under) |
|---|---|
| H2O is commonly called water. | H2O is called water. |

The target appears near the end of the instruction; the matched POS generation is compliant.

### AVOID1 Can Overshoot

**Model:** Claude Haiku 4.5 **Target:** 5 words

**Task:** Name the largest planet in the solar system.

| POS output (count = 5) | AVOID1 output (count = 6, over) |
|---|---|
| Jupiter is the largest planet. | Jupiter is the largest planet here. |

Avoidance wording can fail by overshooting rather than by omitting content.

### AVOID1 Can Also Undershoot

**Model:** Mistral Small **Target:** 5 words

**Task:** What is the square root of 64?

| POS output (count = 5) | AVOID1 output (count = 1, under) |
|---|---|
| Eight times eight equals sixty-four. | 8 |

The same avoidance form can undershoot, showing that the failure direction is not uniform.

### EXCLUDE1 Can Produce an Extreme-Short Output

**Model:** DeepSeek Chat

**Target:** 5 words

**Task:** Name the largest planet in the solar system.

**POS output** count = 5

Jupiter is the largest planet.

**EXCLUDE1 output** count = 1, under

Jupiter

The exclusion construction can trigger answer suppression rather than execution of the count constraint.

### JSON Wrapper Failure Under Strict Full-Output Parsing

**Model:** Claude Haiku 4.5

**Form:** AVOID

**Task:** Name the capital of France.

**Visible output** parse failed

````
```json
{
  "answer": "Paris",
  "reason": "Paris is the capital city of France, located in the north-central part of the
country on the Seine River."
}
```
````

The object is valid-looking, but Markdown fencing makes the complete visible output fail the strict JSON parser.

### JSON Wording Can Improve Strict Compliance

**Model:** GPT-4o mini

**Task:** Hamlet author

**Task:** Who wrote the play Hamlet?

**POS output** parse failed

````
```json
{
  "answer": "William Shakespeare",
  "reason": "William Shakespeare is widely
regarded as the author of Hamlet, which is
considered one of his most famous and
significant works in English literature."
}
```
````

**AVOID output** compliant

```
{
  "answer": "William Shakespeare",
  "reason": "Hamlet is one of the most famous
tragedies written by William Shakespeare,
attributed to his work in the early 17th
century."
}
```

This structural example shows why the claim is directional only within a constraint setting: avoidance wording can reduce wrapper failures for JSON.



\end{document}